%% file: paper.tex
\documentclass[table,dvipsnames]{fairmeta}
\input{math_commands}

\newmdenv[backgroundcolor=metabg, roundcorner=5pt, skipabove=7pt, linewidth=0pt, innertopmargin=4pt]{myframe}

\renewcommand{\eqref}[1]{\labelcref{#1}}

\usepackage{bbding}
\usepackage{amssymb}
\usepackage{booktabs, makecell, tabularx}
\usepackage{tabularray}
\usepackage{tikz}
\usepackage{algorithm}
\usepackage{dsfont}
\usepackage{diagbox}

\title{Set Block Decoding is a Language Model\\ Inference Accelerator}

\author[*]{Itai Gat}
\author[*]{Heli Ben-Hamu}
\author[]{Marton Havasi}
\author[]{Daniel Haziza}
\author[]{Jeremy Reizenstein}
\author[]{Gabriel Synnaeve}
\author[]{David Lopez-Paz}
\author[]{Brian Karrer}
\author[]{Yaron Lipman}

\affiliation[]{FAIR at Meta}

\contribution[*]{Equal contribution}

\abstract{Autoregressive next token prediction language models offer powerful capabilities but face significant challenges in practical deployment due to the high computational and memory costs of inference, particularly during the decoding stage. We introduce Set Block Decoding (SBD), a simple and flexible paradigm that accelerates generation by integrating standard next token prediction (NTP) and masked token prediction (MATP) within a single architecture. SBD allows the model to sample multiple, not necessarily consecutive, future tokens in parallel, a key distinction from previous acceleration methods. This flexibility allows the use of advanced solvers from the discrete diffusion literature, offering significant speedups without sacrificing accuracy. SBD requires no architectural changes or extra training hyperparameters, maintains compatibility with exact KV-caching, and can be implemented by fine-tuning existing next token prediction models. By fine-tuning Llama-3.1 8B and Qwen-3 8B, we demonstrate that SBD enables a 3-5x reduction in the number of forward passes required for generation while achieving same performance as equivalent NTP training.

}

\begin{document}

\maketitle

\section{Introduction}\label{section:intro}\vspace{-10pt}

\begin{wrapfigure}{r}{0.5\textwidth}
\vspace{-0.5cm}
\begin{center}
\includegraphics[width=0.48\textwidth]{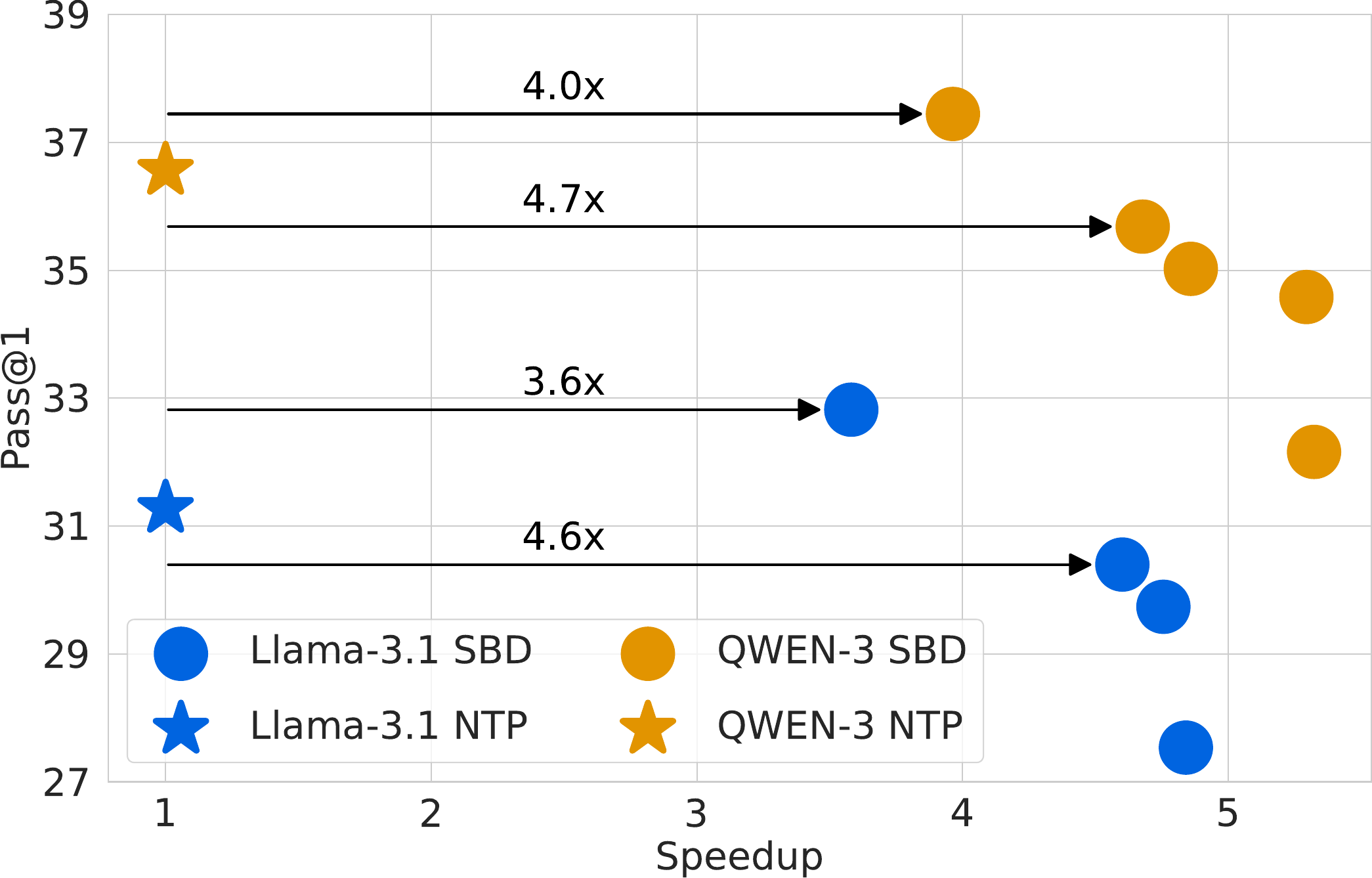}
\end{center}
\caption{LiveCodeBench-V6 acceleration with Set Block Decoding with no performance reductions.}
\vspace{-0.4cm}
\label{fig:speedup_scatters}
\end{wrapfigure}
Next token prediction (NTP) language models (LMs) have demonstrated extraordinary capabilities across a spectrum of tasks, from natural language understanding to complex reasoning and code generation. These models, built upon the Transformer architecture, have scaled to hundreds of billions and even trillions of parameters, a growth that has been directly correlated with their enhanced performance. However, this large scale presents a challenge for practical deployment. The process of using a trained language model to generate predictions, known as inference or sampling, is both computationally expensive and memory-intensive, creating a significant barrier for real-world applications where low latency and high throughput are important.

Language model inference consists of two parts: the \emph{prefilling} and the \emph{decoding} stages. In the \emph{prefilling} stage, the prompt is processed simultaneously and its attention keys and values (KVs) are cached, which usually achieves a high efficiency on GPUs because thousands of tokens are processed together. In the \emph{decoding} stage, tokens are sequentially generated one after the other and require the attention KVs of preceding tokens: this stage requires few FLOPs per token, yet the entire model weights must be read from GPU memory, along with all cached keys and values, as many times as the number of tokens we want to generate.  The decoding stage, which typically dominates the total inference time is the primary focus of many language model optimization efforts. Three main avenues for improvements are: Model compression, system-level optimization, and novel algorithms and modeling. 

One central approach for accelerating language models within the algorithmic/modeling efforts is \emph{speculative decoding}: generate many tokens with a fast \emph{draft} model and then verify and accept them with the slower \emph{target} model. A smaller language model can be used as the draft model \citep{leviathan2023fast}, or additional heads can be attached  to the target language model itself to predict multiple independent tokens into the future \citep{stern2018blockwise,cai2024medusa}. In both cases the target model is used to verify and accept a consecutive subsequence of these tokens. 

The goal of this work is to introduce \emph{Set Block Decoding} (SBD) language models, a flexible and arguably simpler alternative to the draft/target-approach for accelerating language models. SBD models seamlessly combine the standard next token prediction (NTP) paradigm with the masked token prediction (MATP) in the same transformer architecture and, in contrast to previous draft/target-approaches, can sample future tokens in \emph{arbitrary order and in parallel}. This extra degree of freedom unlocks the possibility to employ specialized solvers from the masked diffusion literature, \eg, \citet{ben2025accelerated}, to achieve significant speedups at no accuracy loss and gain a refined control over the speedup-accuracy tradeoff, see \cref{fig:speedup_scatters}. 

Set Block Decoding models offer the following advantages:
\begin{enumerate}
    \item \emph{Simplicity}: Single language model, {no architectural changes and additions; no added hyper-parameters during training; a single new hyper-parameter during inference.} 
    \item \emph{Flexibility}: Can use advanced solvers from the discrete diffusion literature. 
    \item \emph{Efficiency}: Compatible with {exact KV-caching while offering 3-5x fewer model decoding forwards per generated token. }
    \item \emph{Computational-cost-effective}: Can be {quickly fine-tuned from an existing NTP language model.}
\end{enumerate}
Experimentally, we have fine-tuned Llama-3.1 8B and Qwen-3 8B models to make them SBD models and compared to the corresponding NTP baselines trained on the same data and with the same training parameters and noticed that SBD preserve the performance of the NTP models while allows accelerating inference by requiring 3-5x less model forwards.

\begin{figure*}
\centering  
\begin{tabular}{@{}c@{}c@{}}
    \includegraphics[width=0.5\linewidth]{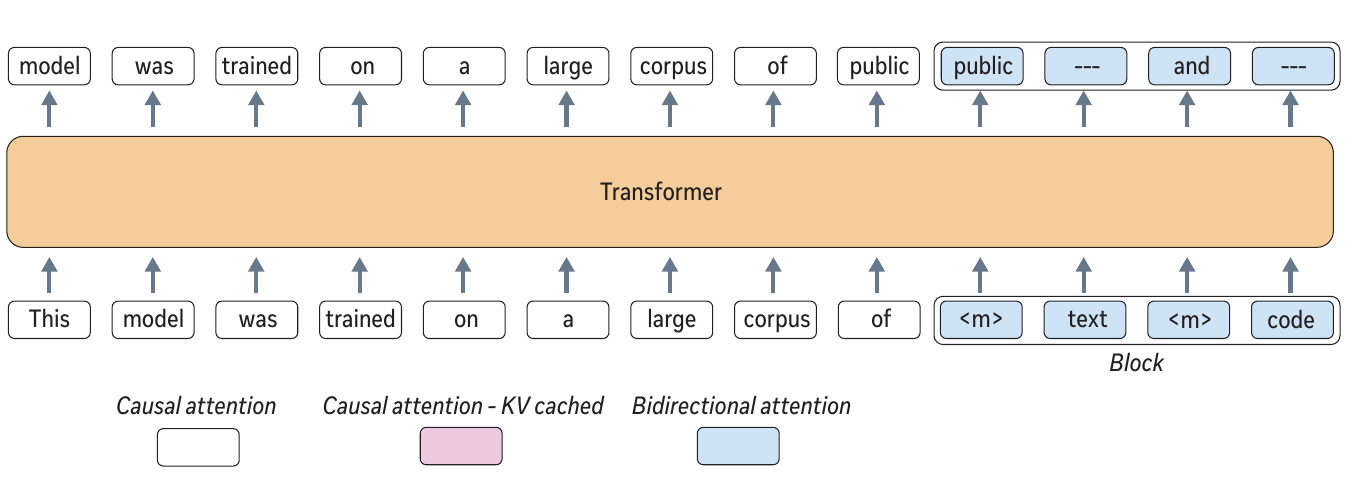}  & \includegraphics[width=0.5\linewidth]{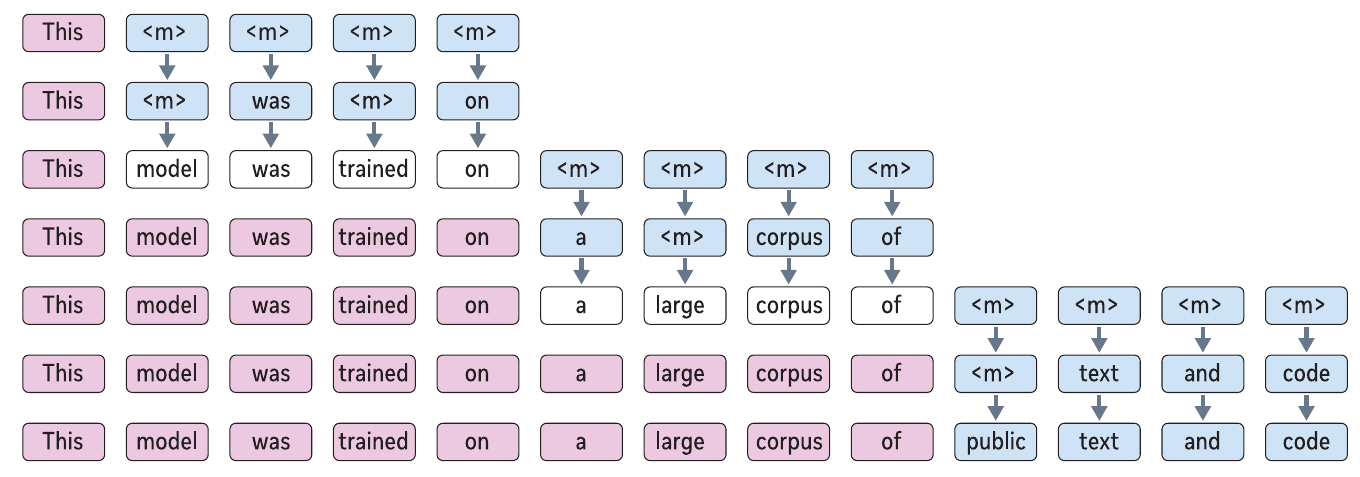} \\
    (a) SBD training/finetuning & (b) SBD inference 
\end{tabular}     
    \caption{(a) \emph{Set Block Decoder} fine-tunes any native NTP transformer architecture to predict $k$ future tokens (in this illustration $k=4$) conditioned on an arbitrary \emph{subset} of the future tokens (in this case, ``text'' and ``code''); where the special mask token `m' is used to hide future tokens to be predicted. Past tokens (in white) use causal attention while the future block (in blue) uses bidirectional attention, allowing future tokens to attend to each other. (b) During inference, SBD decodes one block at a time by revealing some subset of independent future tokens, where each row represents a single forward in the model. Once a block is decoded it is KV-cached (in pink, with causal attention).}
    \label{fig:method}
\end{figure*}

\section{Approach}
In this section we define SBD models as well as how to train and sample them, see \cref{fig:method}. For that end, we start with a bit of background on parallel block decoding. 

\textbf{Notation.} We will mostly follow standard notation denoting a sequence of tokens $x=(x_1,x_2,\ldots,x_L)$, where $L$ is the sequence length and each token is an element of a vocabulary set $x_i\in \gV = \set{1,2,\ldots,V}$. We will use $\gI,\gJ,\gT$ to define sets of indices; accordingly if $\gI=(i_1,\ldots,i_m)$ then $x_\gI=(x_{i_1},\ldots,x_{i_m})$.

\subsection{Parallel block decoding}
Large language models generate a sequence of tokens $x=(x_1,\ldots,x_L)$ by repeatedly predicting the next token $x_t$ given all previous tokens $x_{<t}=(x_1,\ldots,x_{t-2},x_{t-1})$, 
\begin{equation}\label{e:ntp}
    p(x_{t} | x_{<t}).
\end{equation}
This approach is called Next-Token-Prediction (NTP).
NTP inference requires one model evaluation per generated token and is therefore slow. \emph{Parallel block decoding} is an attempt to accelerate language models by making them predict $k$ future tokens (called a \emph{block}) in a single forward pass \citep{stern2018blockwise},
\begin{equation}    p(x_\gI | x_{<t}), \quad \gI=\set{t,t+1,\ldots,t+k-1}
\end{equation}
where $x_\gI = (x_t,x_{t+1},\ldots,x_{t+k-1})$. 
However, since modeling the joint probability of more than a single token is intractable with today's vocabulary sizes $|\gV|$, \eg, the joint probability mass function of two tokens requires $|\gV|^2$ values in general, previous works resorted to \emph{independent parallel decoding}, that is learning only the \emph{marginals}
\begin{equation}
    p(x_{i}|x_{<t}), \text{ for } i\in \gI.
\end{equation}
Independent parallel decoding provides an independent joint for the future tokens,
\begin{equation}    p(x_\gI | x_{<t}) = \prod_{i\in \gI} p(x_{i}|x_{<t}),
\end{equation}
which is in general only a crude approximation to the true joint and therefore requires a verification step that is done with the NTP model and therefore can only  accept some consecutive prefix $x_t,x_{t+1},\ldots,x_{t+a}$. %

\subsection{Set parallel block decoding}
\label{s:masked_parallel_decoding}
A more general parallel decoding framework is the  \emph{set}, or equivalently \emph{mask} parallel decoding \citep{ghazvininejad2019mask,lou2023discrete,gat2024discrete,shi2024simplified,nie2025large}. In a nutshell, Set Block Decoding performs independent parallel decoding of the $k$ future tokens $\gI$ but allows the model to see an arbitrary \emph{subset} $\gJ\subset \gI$ of the future tokens. Equivalently, \emph{mask} some of the $k$ future tokens, $\gM=\gI \setminus \gJ$, and try to predict the masked tokens, 
\begin{equation}\label{e:masked_parallel_decoding}
    p(x_{i} |  x_{<t},x_\gJ), \text{ for } i\in \gM.
\end{equation}
This model provides a lot of flexibility in sampling: for any sequence of indices $\gI_1 \subset \gI_2 \subset \cdots \subset \gI_{\ell}=\gI$ we can decode all the tokens in $\gD_j = \gI_j\setminus \gI_{j-1}$ simultaneously leading to
\begin{equation}
    p(x_\gI | x_{<t}) = \prod_{j=1}^{\ell}  \prod_{i\in \gD_j} p(x_{i}|x_{<t},x_{\gI_{j-1}}).
\end{equation}
This sampling will be exact if all $x_i$, $i \in \gD_j$,  are \emph{conditionally independent}, \ie, 
\begin{equation}    p(x_{\gD_j}|x_{<t}, x_{\gI_{j-1}}) = \prod_{i\in \gD_j} p(x_{i}|x_{<t},x_{\gI_{j-1}}).
\end{equation}
This flexibility was recently explored by \citet{ben2025accelerated} who suggested a practical algorithm utilizing this principle named \emph{Entropy Bounded} (EB) Sampler. In its simplest form, the EB-Sampler finds at each step a subset of tokens to decode in parallel, $\gD_j$, by identifying the masked tokens with low \emph{mutual information}, which quantifies the degree of dependence among these tokens. Since the mutual information cannot be computed without the full joint probability, an upper bound is used instead utilizing the entropy of the marginals $p(x_i|x_{<t},x_{\gJ})$. That is, given we already decoded the tokens corresponding to the indices $\gJ\subset \gI$ and we are left with masks $\gM=\gI\setminus \gJ$, we sort the masked token indices, $\gM$, in ascending order according to entropy of their predicted probabilities, \ie, $H(p(x_{i}|x_{<t},x_\gJ))$; denote the sorted masked indices by $i_1,i_2,\ldots,i_{|\gM|}$. Now, decode in parallel the $s$ tokens $i_1,i_2,\ldots,i_{s}$ where $s\geq 1$ is the largest integer so that 
\begin{equation}\label{e:eb_sampler}
    \sum_{j=1}^{s-1} H(p(x_{i_j}|x_{<t},x_\gJ)) \leq \gamma,
\end{equation}
\input{figures/algorithms}

and $\gamma>0$ is a user prescribed threshold. At least one token is revealed in each iteration of this algorithm. This algorithm, in contrast to the more classical parallel decoding, can control the efficiency-accuracy tradeoff with the single hyperparameter, $\gamma$, and demonstrates impressive reduction in model forwards during sampling. However, translating reductions in model forwards to wall-clock speedups for state-of-the-art autoregressive language models requires combining set parallel decoding with classical next token prediction in a seamless manner. This is what SBD models offer, and this capability is described next.

\subsection{Set block decoding model}
We devise a modeling that allows training of a next token prediction (NTP) model (\cref{e:ntp}) empowered with set block decoding (SBD) abilities (\cref{e:masked_parallel_decoding}). The main benefit of SBD models is that they preserve NTP performance while allowing to accelerate inference by using KV-caching and decoding several tokens simultaneously at each step. 

An \emph{SBD network} is a transformer $f_\theta(\cdot\, ; \, \cdot)$ defined by \begin{myframe}\vspace{-10pt}
\begin{subequations}\label{e:hybrid}
    \begin{align}\label{e:hybrid_causal}
    z_{t} &= f_\theta(x_1,\ldots,x_{t-1} \, ; \, ) \\ \label{e:hybrid_block}
    (\hat{z}_{t},\ldots,\hat{z}_{t+k-1}) &= f_\theta(x_1,\ldots,x_{t-1}\, ; \, \hat{x}_{t},\ldots, \hat{x}_{t+k-1}),
\end{align}
\end{subequations}    
\end{myframe}
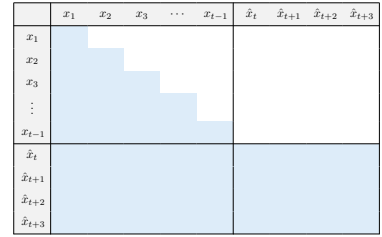
\begin{wrapfigure}[8]{R}{0.35\textwidth}
    \centering
\input{figures/model_attention}

    \caption{The attention mask of the SBD transformer $f_\theta$, \cref{e:hybrid}.}
    \label{fig:model_attention}
\end{wrapfigure}
where all tokens before (left of) `;' are using causal attention, while tokens after (right of) `;' use bidirectional attention, see \cref{fig:model_attention}. The ``noisy'' tokens, $\hat{x}_t,\ldots,\hat{x}_{t+k-1}$, can be either real tokens from $\gV$ or a mask token denoted `m', \ie, $\hat{x}_{i}\in \gV\cup \set{\text{m}}$, for $i\in \gI$; see \cref{fig:method} (a) for an illustration of $f_\theta$ for $k=4$.

As usual, each of the logits $z_t, \hat{z}_t \in\Real^V$ define a probability mass function for the generated token $x_t$ defined via softmax. 
We consequently denote by 
\begin{myframe}\vspace{-10pt}
    \begin{align} \label{e:p_theta_ntp}
        &p_\theta(x_{t}|x_{<t};) \\  \label{e:p_theta_masked}      
        &p_\theta(x_{t+i}|x_{<t}; \hat{x}_{t},\ldots, \hat{x}_{t+k-1}), \quad i = 0,\ldots,k-1 
    \end{align}
\end{myframe}
the probability mass functions defined by the logits $z_t$ (top equation) and $\hat{z}_t, \ldots, \hat{z}_{t+k-1}$ (bottom equation), respectively.

\input{figures/sample_block}

\textbf{Inference with an SBD model.} The SBD model can be sampled with the help of any masked parallel decoding method. We opt for the EB-Sampler where the masked probability in \cref{e:masked_parallel_decoding} is defined using the model as 
\begin{equation}
    p(x_i|x_{<t},x_\gJ) = p_\theta(x_i|x_{<t};\hat{x}_{t},\ldots, \hat{x}_{t+k-1}), \quad i\in\gM,
\end{equation}
and $\gM\subset \gI$ are the masked indices,  $\hat{x}_{j}=x_{j}$ if $j\in \gJ=\gI\setminus \gM$ (\ie, unmasked) and $\hat{x}_j=\text{m}$ otherwise. \Cref{alg:bd_inference} provides a sampling pseudo-code, where $\textsc{sample\_block}$ is the procedure for sampling the next block and is provided in \cref{alg:block_sample}, utilizing the EB-Sampler (\cref{s:masked_parallel_decoding}); see \cref{fig:method} (b) for an illustration of the sampling algorithm with block size $k=4$. 

\newpage
\textbf{Training an SBD model.} Training the SBD model combines next token prediction and masked prediction losses. In more detail, consider given a sequence of tokens $x=(x_1,\ldots,x_L)$, we create an additional masked sequence $\hat{x}=(\hat{x}_1,\ldots,\hat{x}_L)$ by setting, for $i\in [L]$, where $[L]=\set{1,\ldots,L}$,
\begin{equation}\label{e:masked_sequence}
    \hat{x}_i = \begin{cases}
        \text{m} & \text{with prob } \eta \\
        x_i & \text{with prob } 1-\eta
    \end{cases}
\end{equation}
where $\eta$ is randomized uniformly in $(0,1)$. Then the \emph{SBD loss} for the sequence $x$ and block size $k$ is 
\begin{align}\label{e:loss}
    \gL(x,\hat{x};\theta) = -\overbrace{\sum_{t=2}^{L}\log p_\theta(x_t | x_{<t};)}^{\text{Next token prediction}} - \overbrace{ \sum_{t\in \gT} \sum_{i=0}^{k-1} \mathds{1}_{\hat{x}_{t+i}=\text{m}} \log p_\theta(x_{t+i} | x_{<t} ;  \hat{x}_{t},\ldots,\hat{x}_{t+k-1})}^{\text{Masked token prediction}}
\end{align}
where $\mathds{1}_C$ is 1 when condition $C$ holds and otherwise 0, and $\gT = \set{ 1 + \ell k \ \vert \ \ell=0,1,2,\ldots,\lfloor\frac{L}{k}\rfloor-1}$. \Cref{alg:bd_training} provide training pseudo-code, and \cref{fig:method} (a) and \cref{fig:training} illustrate the training (input, target and attention) for the running example of block size $k=4$.

\section{Experiments}

In this section we report experiments and evaluations of the Set Block Decoding method for accelerating large language models. In \cref{sec:benchmarks} we fine-tune two leading 8B models (Llama-3.1 and Qwen-3) and benchmark them on a suite of generation tasks covering reasoning, coding, and mathematics. Next, in \cref{sec:ablation} we investigate the method's scaling properties compared to NTP models on smaller, 3B models. We use these experiments to formulate a training recipe for SBD models. 

\subsection{Benchmarks}\label{sec:benchmarks}

\paragraph{Experimental setup.}
We fine-tune Llama-3.1 8B base~\citep{grattafiori2024llama3herdmodels} and Qwen-3 8B~\citep{yang2025qwen3technicalreport} base models on 70B tokens. We train these models on a mix of reasoning and instruction data with a 32k token context length.  The reasoning data mix is composed of mitigated versions of the OpenCodeReasoning~\citep{ahmad2025opencodereasoning} and OpenMathReasoning~\citep{moshkov2025aimo2} datasets, where mitigations including algorithmic bias filtering and cybersecurity protections, were applied. For instruct data, we use a publicly available mix, similar to the one used for training Llama-3.1 Instruct.
For optimization, we use AdamW~\citep{loshchilov2019decoupledweightdecayregularization} with a 3e-4 learning rate, a warmup of 200 iterations, and cosine annealing schedule. We use a batch size of 2M tokens and total 34k iterations of fine-tuning. To support our method, we train models with a variable block size by uniformly sampling a size from the range [2,16] at each training step. At inference, we use EB-Sampler as described in \cref{e:eb_sampler} with temperature 0.

\paragraph{Results.} \Cref{tab:thinking_generation_benchmarks} presents results for our SBD models on reasoning benchmarks  AIME25, LiveCodeBench v6~\citep{jain2024livecodebench}, Math500~\citep{lightman2023lets}, as well as chat benchmarks GSM8K~\citep{cobbe2021trainingverifierssolvemath}, HumanEval+~\citep{evalplus}, and MBPP~\citep{austin2021program}. For reasoning benchmarks, we generate with thinking and up to 32k tokens; for chat benchmarks we do not use thinking and generate 1024 tokens for HumanEval+ and GSM8K, and 256 for MBPP. The $\gamma$ values we used for reasoning are  $\gamma_{\text{low}}=0.1,\gamma_{high}=0.35$; while for chat benchmarks we use $\gamma_{\text{low}}=0.35,\gamma_{high}=0.6$. 

To isolate the effect of training data quality on performance and to ensure a fair comparison between SBD and NTP, for each base model we trained an NTP baseline, corresponding to NTP in the loss column in \cref{tab:thinking_generation_benchmarks}, in the exact same experimental setting and confirmed that the SBD performance is consistent with it.

\pagebreak
We find that for low $\gamma$ values, our method preserves the autoregressive performance while achieving a 3-5x speedup (benchmark-dependent) and higher $\gamma$ values yield even greater speedups at the cost of a minor drop in performance. \Cref{fig:speedup_scatters} illustrates the tradeoff between speed and performance when tweaking the $\gamma$ threshold of the EB-Sampler.

\emph{In all benchmarks, similar to the ``generate until'' logic presented in~\citet{ben2025accelerated}, we measure speedup only until the end of the problem's solution.}

\begin{table}[t]
  \centering
    \resizebox{\textwidth}{!}{%
    \begin{tabular}{llllllllllll}
    \toprule
    \multirow{2}{*}{Base Model} & \multirow{2}{*}{Training} & \multirow{2}{*}{Loss} & \multirow{2}{*}{Sampling}  & \multirow{2}{*}{Speedup}  & \multicolumn{3}{c}{Reasoning}  & \multicolumn{3}{c}{Chat} \\
    \cmidrule(r){6-8} \cmidrule(l){9-11}
    & & & &   & MATH500 & AIME25 & LCB V6  & GSM8K & HE+ & MBPP \\
    \midrule
    Gemini diffusion & - &  - & Diffusion & - & - & 23.3 & 30.9  &  - & -& 76.0 \\
    Mercury & - & - & Diffusion &- & -  & - & 25.0 &  - & -  & 76.6 \\
    Diffucoder & Scratch & MATP & Diffusion & 1x & - & - & -  &  - & 68.3 & 67.5 \\
    Llada 1.5 & Scratch &  MATP &Diffusion & 1x & - & -  & -  &  83.3 & 52.4 & 42.8\\
    Dream-coder & FT & MATP & Diffusion & 1x & - & - & 21.4  & -& - & 79.6 \\
    \midrule
    \multirow{4}{*}{Llama-3.1 8B} & \multirow{4}{*}{FT}  & NTP & NTP& 1x & 80.2 & 33.3 & 31.5 & 85.3 & 56.7 & 70.4\\
    & & SBD & NTP & 1x  & 81.6 & 30.0 & 31.3 & 85.6 & 57.9 & 70.9\\
    & & SBD & SBD ($\gamma_{\text{\tiny high}}$)& 3.4x  & 80.4~\xr{3.23} & 23.3~\xr{4.50} & 29.9~\xr{4.59}	& 84.0~\xr{2.34} & 54.9~\xr{2.95} & 67.2~\xr{2.66}\\
    & & SBD & SBD ($\gamma_{\text{\tiny low}}$)& 3.0x  & 81.0~\xr{3.55} & 30.0~\xr{3.35} & 31.7~\xr{3.72} & 84.2~\xr{2.20} & 57.9~\xr{2.61} & 69.5~\xr{2.49} \\
    \midrule
    \multirow{4}{*}{Qwen-3 8B}  & \multirow{4}{*}{FT} & NTP & NTP & 1x & 86.6 & 33.3 & 36.6 & 90.1 & 69.5 & 78.0 \\
    & & SBD & NTP &  1x & 86.6 &  33.3& 37.4 & 90.4 & 66.5 & 77.7\\
    & & SBD & SBD ($\gamma_{\text{\tiny high}}$) & 3.8x & 85.4~\xr{3.54} & 26.6~\xr{5.06} & 33.3~\xr{5.36} & 88.7~\xr{2.86} & 65.2~\xr{2.72} & 74.6~\xr{3.03} \\
    & & SBD & SBD ($\gamma_{\text{\tiny low}}$) & 3.2x & 85.0~\xr{3.41} & 33.3~\xr{3.85} & 37.2~\xr{3.92} & 90.1~\xr{2.77} & 66.5~\xr{2.51} & 77.5~\xr{2.61}\\
    \bottomrule
    \end{tabular}%
    }
    \caption{Reasoning, coding, and math benchmark results. All baselines use greedy decoding. SBD models are sampled with a fixed entropy threshold and two $\gamma$ settings: a low value that preserves accuracy and a high value that prioritizes speed. Speedup is measured as the reduction in NFEs (model forwards); see \cref{sec:roofline} for the wall-clock time analysis.\vspace{-0pt}}
    \label{tab:thinking_generation_benchmarks}
\end{table}

\subsection{Ablations}\label{sec:ablation}

\paragraph{Experimental setup.} For all experiments in this section we use a 3B transformer model, with 28 layers, and a hidden dimension of 3072. For pretraining experiments, we use  AdamW with a peak learning rate of 1.5e-3, warmup of 2000 steps and a cosine annealing schedule, for a total of 1T tokens from a mitigated version of DCLM~\citep{li2024datacomplm} and raw code data. For instruct SBD fine-tuning, we use AdamW with a peak learning rate of 1e-5, warmup of 200 steps and a cosine annealing schedule. We train with the same instruct data used in \cref{sec:benchmarks}.

\begin{wrapfigure}{r}{0.4\textwidth}
\vspace{-0.45cm}
\begin{center}
\includegraphics[width=0.38\textwidth]{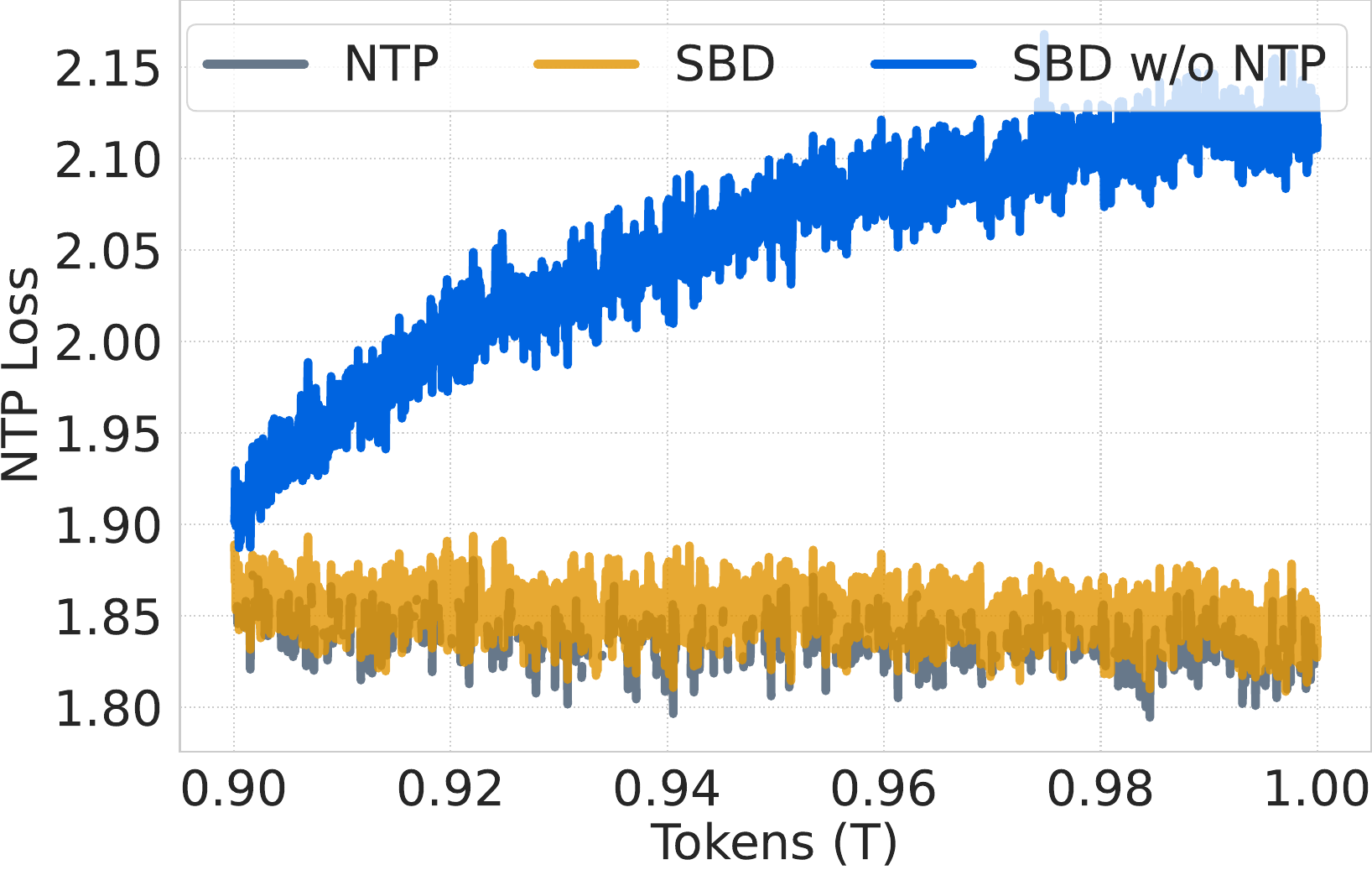}
\end{center}
\caption{NTP loss during 3B model pretraining.}
\label{fig:role_ntp_loss}
\vspace{-0.45cm}
\end{wrapfigure}

\paragraph{Role of NTP loss term.} We assess the significance of the NTP loss term (see \cref{e:loss}) by training two variants: standard SBD, and a variant where we do not take gradients over the NTP term, but we do log its values during training. Both variants begin from the same intermediate training step, namely the 900B checkpoint of a standard NTP model pretraining. We train SBD with and without the NTP term for the remaining 100B tokens, continuing with the same exact optimization hyperparameters as the NTP training.

\Cref{fig:role_ntp_loss} shows the NTP loss during training in three configurations: (i) standard NTP model training (gray), (ii) SBD training (orange), and (iii) SBD training w/o NTP term (blue). One can see that without the NTP loss term, SBD training will not maintain the AR capabilities of the base model, while training SBD with the full loss only results in a slightly higher NTP loss.

\begin{table}[H]
  \centering
  \footnotesize
  \begin{tabular}{lllllll}
    \toprule
    & MMLU & GPQA & Hellaswag & Winogrande & ARC-E & ARC-C \\
    \midrule
    NTP & 50.9 & 24.1 & 76.4 & 69.5 & 71.2 & 51.4   \\
    SBD  & 49.9 \mb{1.0} & 27.7 \pb{3.6} & 75.6 \mb{0.8} & 70.2 \pb{0.7} & 69.9 \mb{1.3} & 48.6 \mb{2.8}  \\
    SBD \textsubscript{w/o NTP loss}   & 43.2 \mr{7.7} & 27.2 \pb{3.1} & 71.1 \mr{5.2} & 67.2 \mb{2.2} & 58.8 \mr{12.4} & 43.2 \mr{8.2}  \\
    \bottomrule
  \end{tabular}%
  \caption{Ablation on 3B-parameters pretrained models trained on 1T tokens. SBD models start from an intermediate training step of the NTP pretrained model and are trained for the last $10\%$ of the tokens. The difference in accuracy compared to the NTP baseline is listed in subscript for the SBD models.\vspace{-10pt}}
  \label{tab:3B_ablation_ntp_loss}
\end{table}

\Cref{tab:3B_ablation_ntp_loss} further validates the observed behavior in \cref{fig:role_ntp_loss} by evaluating the models on likelihood tasks: MMLU \citep{hendryckstest2021}, GPQA \citep{rein2023gpqa}, Hellaswag \citep{zellers2019hellaswag}, Winogrande \citep{sakaguchi2019winogrande}, Arc-E and ARC-C \citep{Clark2018ThinkYH}, showing significant decrease in accuracy when the NTP loss term is removed compared to full SBD training loss. For the SBD models, inference is done using the NTP prediction as in \cref{e:hybrid_causal}.

\begin{figure}[t]
  \centering
  \begin{tabular}{ccc}
     \footnotesize HumanEval  & \footnotesize MBPP &
     \footnotesize GSM8K \\
     \includegraphics[width=0.3\textwidth]{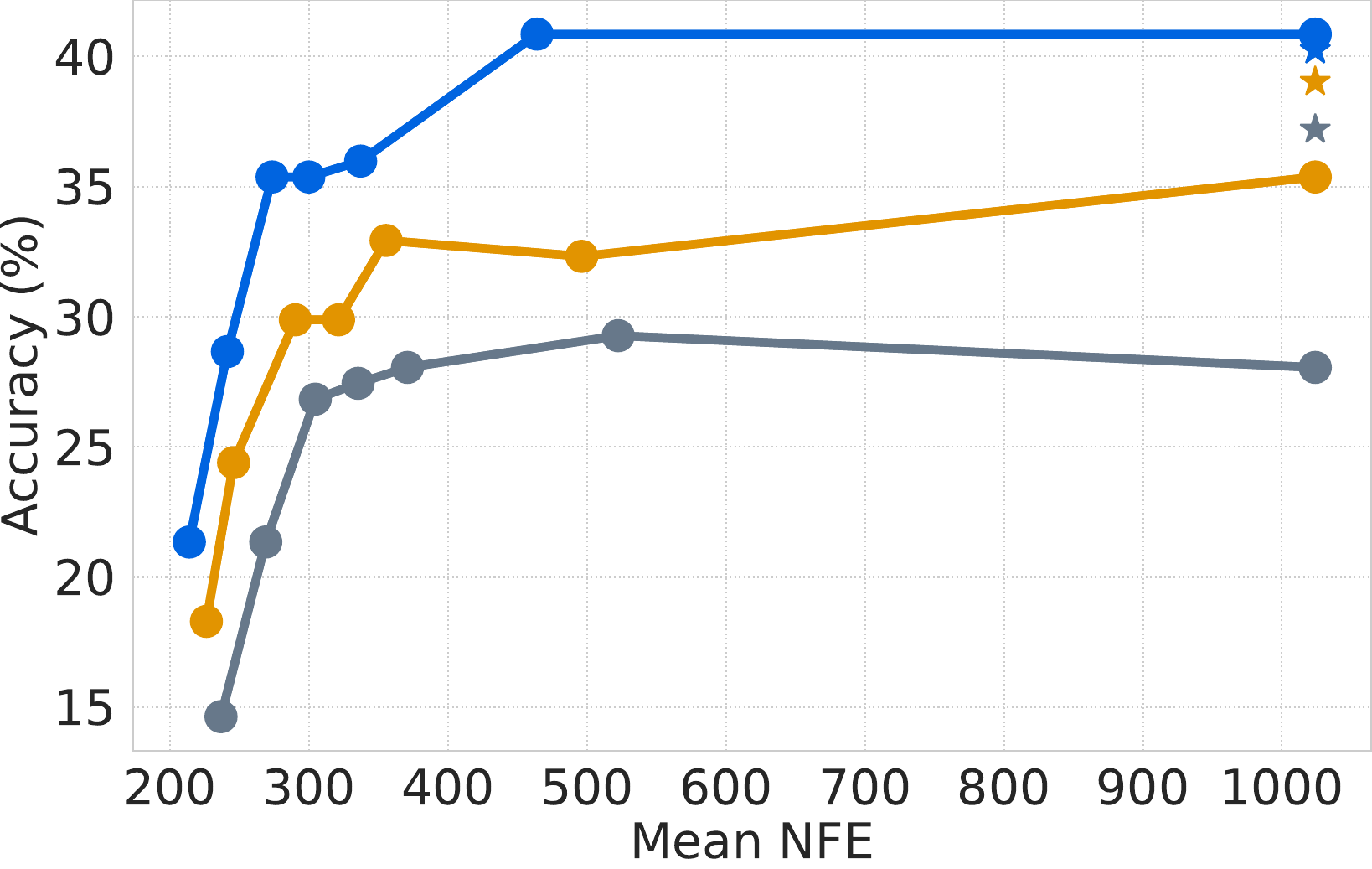}  & \includegraphics[width=0.3\textwidth]{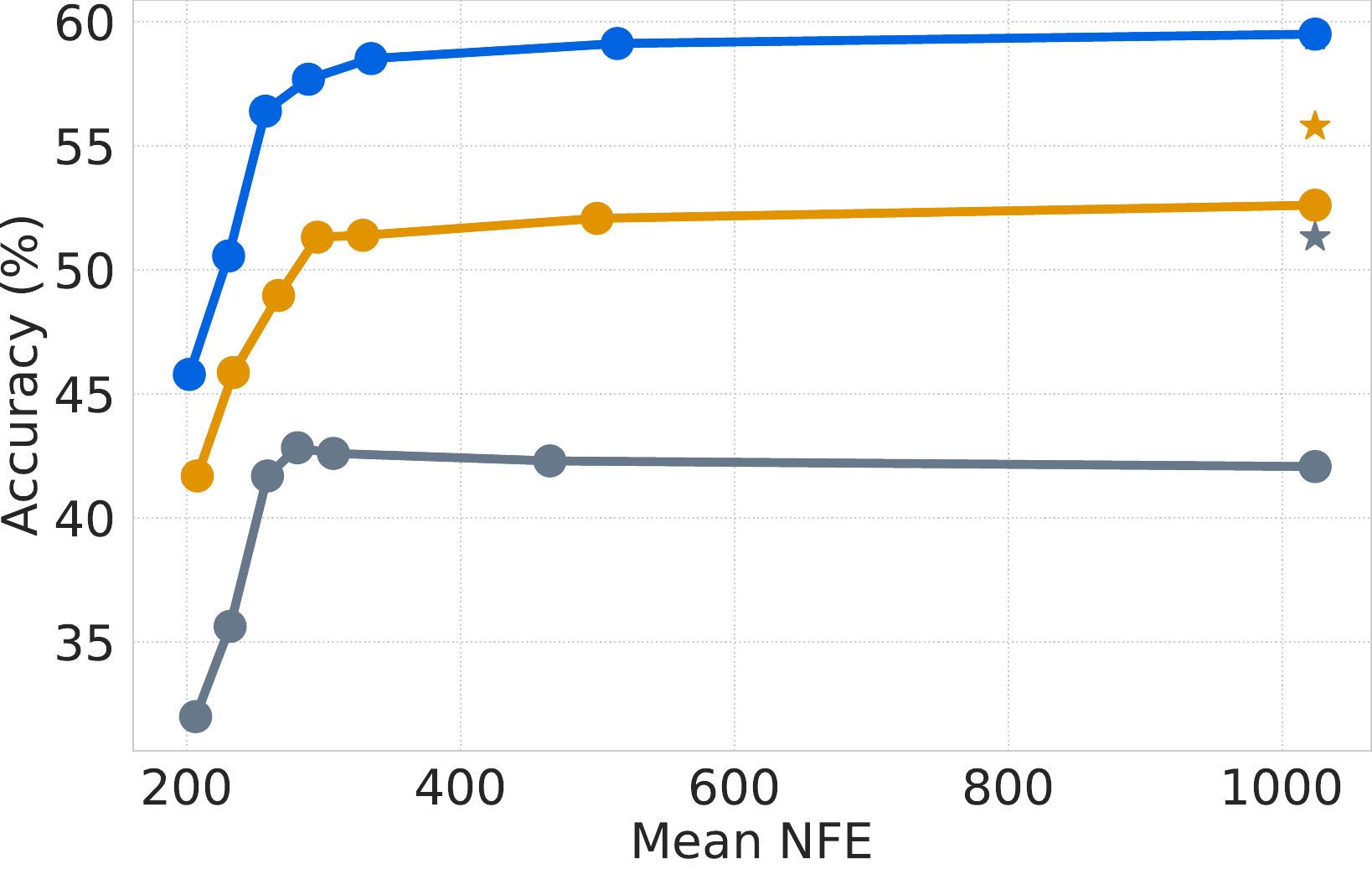} &     \includegraphics[width=0.3\textwidth]{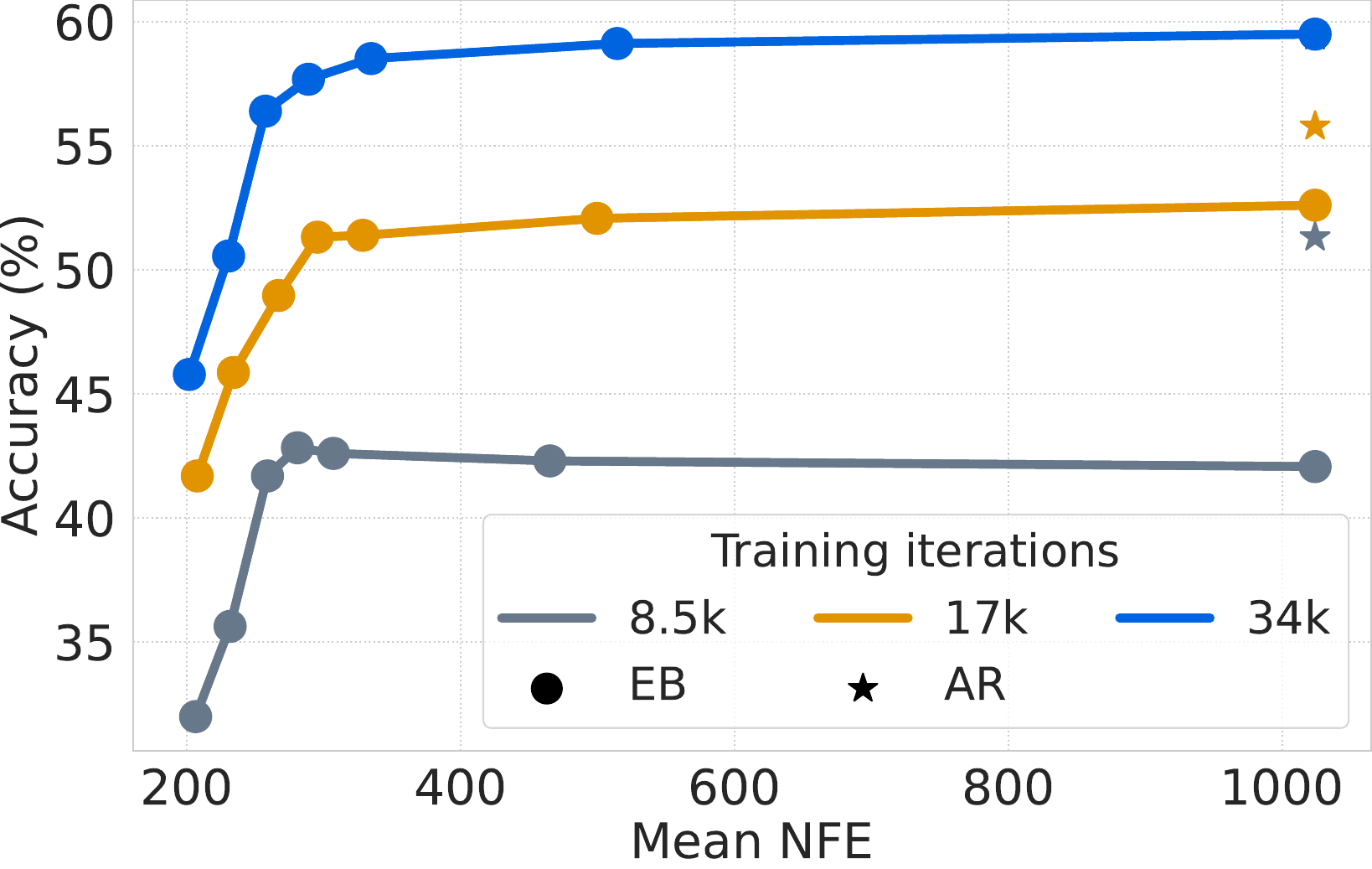}
  \end{tabular}
  \vspace{-0.2cm}
  \caption{3B model SFT training with varying number of training iterations.}
  \label{fig:3B_sft}
\end{figure}

\paragraph{Number of training steps.} State-of-the-art language models are first fully pretrained and then undergo a supervised fine-tuning (SFT) in order to adjust them for specific use cases (\eg, instruct models). The SFT stage is typically significantly shorter than the full pretraining stage and is one of the most 
common post-training procedures. In this experiment we show SBD can be incorporated effectively into the SFT stage. That is, given a pretrained model, perform SFT with the SBD training scheme. We use the same instruct data used in \cref{sec:benchmarks} and the same training hyperparameters for both NTP and SBD fine-tuning.  \Cref{fig:3B_sft} shows SFT models' performance for both NTP training and SBD training at 1024 generation length, for varying number of training iterations. We observe that SBD requires more steps to reach on-par performance as NTP training but closes the performance gap after roughly 34k training iterations. We also show the performance of the SBD models for different EB-Sampler $\gamma$ values in $\{0,0.01,0.1,0.2,0.4,0.8,1.5 \}$. This demonstrates that the EB-Sampler's behavior, providing speed up at no accuracy loss, as shown in \cite{ben2025accelerated} for masked models, translates to the SBD paradigm. In \cref{app:factor} we also show comparison to the factor sampling algorithm introduced in \cite{wu2025fast}.

\pagebreak

\section{Timing analysis}\label{sec:roofline}

\vspace{-0.2cm}\subsection{Roofline model for block inference}\label{sec:roofline_model}
Standard NTP models perform $L$ forwards to generate a sequence of length $L$. In contrast, SBD models reduce this number of model forwards by a factor of 3-5x (see factors in \cref{tab:thinking_generation_benchmarks}) but each forward simultaneously computes a block of $k$ tokens. In this section we provide a theoretical (``roofline'') analysis to justify why these model forward saving are expected to translate to almost identical factors of wall-clock savings in practice.

Our model assumes the H100 Nvidia GPU and standard 8B transformer model with the following parameters:\vspace{-15pt}
\begin{figure}[H]
  \centering
\begin{minted}{python}
# --- H100 GPU Specifications ---
BYTES: int = 1  # model weights, we assume FP8
PEAK_FLOPS: float = 1978 * 1e12  # peak flops for FP8
PEAK_FLOPS_ATTENTION: float = 989 * 1e12  # attention is done in BF16
MEMORY_BANDWIDTH: float = 3.35 * (1024**4)  # memory bandwidth in B/s

# --- Transformer Model Configuration ---
HIDDEN_DIM: int = 4096
FFN_DIM: int = 4 * HIDDEN_DIM
NUM_HEADS: int = 32
HEAD_DIM: int = HIDDEN_DIM // NUM_HEADS
NUM_KV_HEADS: int = 8
BYTES_PER_KV_ELEMENT: float = 1  # bytes used per KV element
\end{minted}\vspace{-25pt}
  \caption{Constants for roofline analysis.}  \label{code:constant_roofline}
\end{figure}
The \texttt{PEAK\_FLOPS} defines the maximal flops/sec for \texttt{FP8} in H100 while \texttt{MEMORY\_BANDWIDTH} is the maximal bytes/sec. 
The main equation to determine the theoretical time of an operation is to take the max upon memory transfers time and compute time via
\begin{equation}\label{e:theoretical_times}    \texttt{theoretical\_time} = \max\set{\frac{\texttt{total\_flops}}{\texttt{PEAK\_FLOPS}}, \frac{\texttt{total\_memory}}{\texttt{MEMORY\_BANDWIDTH}}}.
\end{equation}
If the maximum is achieved by the memory part then the computation is called \emph{memory bound} and otherwise it is named  \emph{compute bound} \citep{williams2008roofline}.

To compare standard NTP and SBD decoding time we analyze different KV Cache lengths, supporting different stages of the generation process with potentially long generation sequences, different sizes of block sizes $\set{1,2,4,8,16,32,64}$, where block size of 1 corresponds to NTP sampling and serves as baseline, and batch size, $\texttt{batch\_size}\in\set{1,4,8,16}$. To analyze the theoretical forward time of a transformer we sum the theoretical times of its two components, namely the multi-head attention operation and linear layer. For the attention, we assume a single fused kernel like in \cite{dao2022flashattention}. In the appendix, we provide Python code to calculate these theoretical runtimes (\cref{code:transformer_roofline}). \Cref{tab:roofline_ours_vs_ar} shows the slowdowns, \ie, the theoretical time ratios, of block sampling and NTP sampling for different block and batch sizes and KV cache length. For example, for block size of 16, the block inference slowdown over NTP is not more than few percents. In the next section, we use this analysis to provide theoretical wall-clock saving times for different NFE speedups.

\begin{table}[htbp]
\centering
\begin{tabular}{cc}
  \resizebox{0.48\textwidth}{!}{\input{figures/roofline/tables/slowdown_table_batch_1} }  &  
  \resizebox{0.48\textwidth}{!}{\input{figures/roofline/tables/slowdown_table_batch_4} }\\
  {\footnotesize \texttt{batch\_size}=1} & {\footnotesize \texttt{batch\_size}=4} \\[5pt]
\resizebox{0.48\textwidth}{!}{\input{figures/roofline/tables/slowdown_table_batch_8} }  &  
  \resizebox{0.48\textwidth}{!}{\input{figures/roofline/tables/slowdown_table_batch_16} } \\
  {\footnotesize \texttt{batch\_size}=8} & {\footnotesize \texttt{batch\_size}=16}
\end{tabular}
\caption{Slowdown factors of a single block forward (for different batch and block sizes and KV cache lengths) compared to a single token forward used in standard NTP sampling, according to theoretical ``Roofline'' analysis for an H100 NVIDIA GPU and a standard 8B transformer architecture. Uncolored cells exhibit small slowdown. \textit{Note: no NFE speedup is considered here.}}\label{tab:roofline_ours_vs_ar}
\end{table}

\begin{table}[t]
\centering
\begin{tabular}{@{}c@{\hskip 5pt}c@{\hskip 5pt}c@{}}
  \resizebox{0.32\textwidth}{!}{\input{figures/roofline/tables/speedup_table_batch_1_nfe_speedup_2} }  &    
  \resizebox{0.32\textwidth}{!}  {\input{figures/roofline/tables/speedup_table_batch_4_nfe_speedup_2} }  &    
  \resizebox{0.32\textwidth}{!}{\input{figures/roofline/tables/speedup_table_batch_8_nfe_speedup_2} }    
  \\
  {\footnotesize \texttt{batch\_size}=1, \texttt{NFE\_speedup}=2}
  & {\footnotesize \texttt{batch\_size}=4, \texttt{NFE\_speedup}=2}
  & {\footnotesize \texttt{batch\_size}=8, \texttt{NFE\_speedup}=2} \\[5pt]
  \resizebox{0.32\textwidth}{!}{\input{figures/roofline/tables/speedup_table_batch_1_nfe_speedup_4} }  &    
  \resizebox{0.32\textwidth}{!}  {\input{figures/roofline/tables/speedup_table_batch_4_nfe_speedup_4} }  &    
  \resizebox{0.32\textwidth}{!}{\input{figures/roofline/tables/speedup_table_batch_8_nfe_speedup_4} }    
  \\
  {\footnotesize \texttt{batch\_size}=1, \texttt{NFE\_speedup}=4}
  & {\footnotesize \texttt{batch\_size}=4, \texttt{NFE\_speedup}=4}
  & {\footnotesize \texttt{batch\_size}=8, \texttt{NFE\_speedup}=4} \\[5pt]
  \resizebox{0.32\textwidth}{!}{\input{figures/roofline/tables/speedup_table_batch_1_nfe_speedup_8} }  &    
  \resizebox{0.32\textwidth}{!}  {\input{figures/roofline/tables/speedup_table_batch_4_nfe_speedup_8} }  &    
  \resizebox{0.32\textwidth}{!}{\input{figures/roofline/tables/speedup_table_batch_8_nfe_speedup_8} }    
  \\
  {\footnotesize \texttt{batch\_size}=1, \texttt{NFE\_speedup}=8}
  & {\footnotesize \texttt{batch\_size}=4, \texttt{NFE\_speedup}=8}
  & {\footnotesize \texttt{batch\_size}=8, \texttt{NFE\_speedup}=8} 
\end{tabular}
\vspace{-0.2cm}
\caption{Theoretical wall-clock speedups of block decoding compared to standard NTP sampling for different NFE speedups based on roofline analysis. Uncolored cells exhibit wall-clock speedup roughly equivalent to NFE speedup. }
  \label{tab:wallclock_block_vs_ar}
  \vspace{-0.2cm}
\end{table}

\subsection{Set block decoding theoretical speedups}

Consider sampling a block of $k$ tokens from the model. We want to estimate the wall-clock speedup of SBD sampling over NTP. Recall the SBD inference procedure in \cref{alg:bd_inference}, which is illustrated in \cref{fig:method} (b), and assume that the SBD sampling uses $l$ model forwards ($l < k$). Then the wall-clock speedup will be the ratio  
\begin{equation} \texttt{speedup}(l,k) = \frac{k\,\text{time(1)}}{\text{time(2$k$)} + (l-1) \text{time($k$)}},   
\end{equation}
where time($k$) is the forward time of a block of size $k$ in the network.

To estimates these times, we will use the roofline analysis from \cref{sec:roofline}. \Cref{tab:wallclock_block_vs_ar} depicts \texttt{speedup}$(l,k)$ for different batch and block sizes, KV cache length and 
\begin{equation}
    \texttt{NFE\_speedup} = \frac{k}{l}.
\end{equation}
Note that for block size 16 the NFE speedups translate almost directly to theoretical wall-clock speedups, which covers the experimental setting and provides an evidence for potential 3-5x wall-clock speedup given the NFE speedups in \cref{tab:thinking_generation_benchmarks}. Lastly, we note that as batch sizes and/or block size increase the roofline analysis indicates diminishing returns due to increased computational costs of block forwards (see colored cells).

\vspace{-0.2cm}\section{Related work}\vspace{-0.2cm}

Efficient large language modeling is a large topic that we do not attempt to cover comprehensively. Instead we summarize research that is most connected to SBD, describing recent diffusion language models as well as related efficiency efforts. Like SBD, this research proposes higher efficiency via computationally cheaper and fewer model evaluations, e.g., applying key-value caching, fusing sequential operations as in tree attention, and parallel decoding.  We conclude by discussing hybrid language models.

\vspace{-0.2cm}\paragraph{Diffusion language models using masked discrete diffusion.}  Large language modeling via diffusion at the several billion (or larger) parameter scale has only recently become successful with models such as Dream \citep{Dreamon2025, dreamcoder2025}, LLaDa \citep{you2025llada, liu2025longllada}, MMaDa \citep{yang2025mmada}, Dimple \citep{yu2025dimple}, Mercury \citep{labs2025mercury}, DiffuCoder \citep{gong2025diffucoder}, Seed Diffusion \citep{song2025seeddiffusion}, and Gemini Diffusion \citep{gemini2025}.  Prior to the arrival of these masked diffusion models, discrete diffusion for text had been mostly limited to smaller scales. These masked diffusion models are competitive with traditional autoregressive language models on task performance, but are still lacking on end-to-end efficiency, outside commercially developed models, such as Mercury, Seed Diffusion, and Gemini Diffusion, whose inner workings are undisclosed.  

\vspace{-0.2cm}\paragraph{Efficient large language models with causal attention.} Typical language models are autoregressive with a fixed left-to-right order and leverage Transformers using causal attention, enabling the reuse of model computation via key-value caching.  This caching is crucial for efficient sampling from these models.  Greedy decoding \citep{stern2018blockwise} and non-greedy speculative sampling \citep{chen2023accelerating, leviathan2023fast} can improve efficiency further, by proposing candidate sequences from a cheap draft model and only evaluating those candidates with the NTP language model.  For example, text diffusion was proposed as a draft in \citet{christopher2025}.  Because multiple models adds system complexity, approaches such as multi-token prediction \citep{gloeckle2024better}, Medusa \citep{cai2024medusa}, and Eagle \citep{li2024eagle}, add output heads to an existing language model to predict consecutive future tokens.  This multi-head prediction enables self-speculative decoding without a separate draft. Unlike our approach, these methods require adding architectural constructions which introduce a tradeoff challenge and a large to explore hyperparameter space.
Any-order autoregressive models \citep{uria14, hoogeboom2022autoregressive} discards the left-to-right order but maintains causal attention and exact key-value caches, and have similar draft variants \citep{pannatier2024sigma, guo2025reviving}. 

\vspace{-0.2cm}\paragraph{Efficient diffusion language models with bidirectional attention.}  As strongly performing masked diffusion models are recent, efficient sampling is even more recent and has focused on improvements to LLaDa and Dream.  This research has introduced approximate key-value caching \citep{ma2025dkv, liu2025dllmcache} and adaptive multi-token sampling \citep{ben2025accelerated}, and explored both avenues simultaneously \citep{wu2025fast, hu2025accelerating, israel2025accelerating}.  Exact key-value caching is not possible with bidirectional attention, and heuristic approximations rely upon utilize empirical observations of slowly changing representations, especially for mask tokens.  On the sampling side, \citep{ben2025accelerated, wu2025fast} propose non-greedy and greedy decoding schemes designed to control error from parallel multi-token sampling, while \citet{hu2025accelerating, israel2025accelerating} consider leveraging autoregressive model outputs to correct for independent unmasking.  Unmasking orders are often constrained, with semi-autoregressive blockwise decoding used in \citet{wu2025fast, hu2025accelerating} and even left-to-right decoding suggested~\citep{israel2025accelerating}.

\vspace{-0.2cm}\paragraph{Hybrid language models.} Most related to SBD  are recent proposals to mix left-to-right and parallel modeling. Block diffusion (BD3-LM) \citep{arriola2025block}, proposes semi-autoregressive generation within blocks, uses block-causal attention and exact key-value caching for preceding blocks, proposes unmasking tokens within a block using diffusion and bidirectional attention. CtrlDiff built upon BD3-LM adding adaptive block size selection \citep{huang2025ctrldiff}.  \citet{fathi2025unifying} concurrently built upon BD3-LM to combine NTP and block diffusion, however with the goal of testing hybrid probability paths with a mix of uniform and masked noising process for improving performance at the cost of longer inference.
Finally Esoteric Language Models \citep{sahoo2025esoteric}, claim to improve upon BD3-LM, by considering a hybrid construction that uses bidirectional attention over clean tokens and causal attention over masked tokens to enable KV-caching. While these methods focus on a similar goal, fusing NTP with MATP models, our work introduces an efficient method to fine-tune an existing NTP model, taking advantage of the efficient NTP training, while providing it with the ability for fast block decoding; this allows us to gain a 3-5x speedup without altering the model architecture or compromising its performance.

\vspace{-0.2cm}\section{Conclusion and future work}\vspace{-0.2cm}

This work introduces Set Block Decoding (SBD), a simple and effective paradigm for accelerating the inference of large language models. By integrating masked-token prediction directly into a standard autoregressive architecture, SBD models can decode multiple, non-consecutive tokens in parallel. SBD avoids the complexity of auxiliary models and complex architectural constructions and requires no architectural changes, making it a practical solution that can be readily implemented by fine-tuning existing language models. Our experiments with Llama-3.1 8B and Qwen-3 8B demonstrate that SBD reduces the number of required forward passes by 3x-5x without compromising the model's original performance.

There are several interesting directions for future work. A key direction is scaling SBD to even larger models to investigate its scaling properties. Furthermore, developing hardware-aware inference implementations to match the theoretical roofline analysis, as well as exploring a wider range of advanced samplers from the discrete diffusion literature, could further unlock the potential of SBD to maximize wall-clock speedups.

\section{Acknowledgements}\vspace{-0.2cm}

We thank Shimon Nowik for his contributions to the method's visualizations and Grigory Sizov for his support with the practical implementation.

\clearpage
\newpage
\bibliographystyle{assets/plainnat}
\bibliography{paper}

\clearpage
\newpage
\beginappendix

\section{Implementation}

\subsection{Method}

The following PyTorch snippets highlight the key modifications for our method, focusing on the changes to a standard autoregressive framework. Code Block~\ref{app:code:train_code} presents the adjustments to the training loop required to learn a hybrid model. The subsequent code blocks detail our custom attention mechanism, implemented using FlexAttention~\citep{dong2024flexattentionprogrammingmodel}: code block~\ref{app:code:train_attention_code} for training and code block~\ref{app:code:inference_attention_code} for inference.

\begin{figure}[H]
  \centering
\begin{minted}{python}
input_ids = sequence[:, :-1]

ar_label_ids = sequence[:, 1:]
parallel_label_ids = input_ids.clone()

bsz, seq_len = input_ids.shape

block_len = torch.randint(max_block_size, (1, )).item()
attention_mask = create_attention_mask_train(seq_len=seq_len, block_len=block_len)

# 1. Compute masked_input
time = torch.rand(size=(bsz, 1), device="cuda")
input_ids_mask = torch.rand(size=input_ids.shape, device=input_ids.device) > time

masked_input = torch.where(condition=input_ids_mask, input=tokenizer.mask_id, other=input_ids)

input_ids = torch.cat([input_ids, masked_input], dim=1)

# 2. Compute label_ids
parallel_label_ids[:, :-1][input_ids_mask] = -100

label_ids = torch.cat([ar_label_ids, parallel_label_ids], dim=1)

# 3. Each unique token location (in AR and parallel) should get the same positional embeddings.
positional_embeddings = get_positional_embeddings(seq_len=seq_len)
positional_embeddings = positional_embeddings.repeat(2, 1)

loss = model(input_ids, mask=attention_mask, targets=label_ids, positional_embeddings=positional_embeddings)
loss.backward()

...
\end{minted}
  \caption{The required modifications to the training loop of an autoregressive model. We first change how the attention mask is computed, then mask a portion of the input tokens. The block-length is chosen randomly.}
  \label{app:code:train_code}
\end{figure}

\clearpage

\input{figures/training_diagram}

\begin{figure}[H]
  \centering
\begin{minted}{python}
from torch.nn.attention.flex_attention import BlockMask, create_block_mask

def create_attention_mask_train(seq_len: int, block_len: int) -> BlockMask:  
    half_seq_len: int = seq_len // 2
    
    def mask_mod(b, h, q_idx, kv_idx):
        # Top-left quadrant (x1 -> x1, standard causal)
        # True if query and key are in the first half and key is before or at query.
        is_in_top_left_causal = (q_idx < half_seq_len) & (kv_idx < half_seq_len) & (kv_idx <= q_idx)
    
        # Bottom-right quadrant (xt -> xt, block attention)
        # True if query and key are in the second half and belong to the same block.
        q_block_xt = (q_idx - half_seq_len) // block_len
        kv_block_xt = (kv_idx - half_seq_len) // block_len
        is_in_bottom_right_block = (q_idx >= half_seq_len) & (kv_idx >= half_seq_len) & (q_block_xt == kv_block_xt)
    
        # Bottom-left quadrant (xt -> x1, block causal past)
        # True if query is in the second half, key is in the first half, and the queries block index is strictly greater than the keys block index.
        q_block_idx_bl = (q_idx - half_seq_len) // block_len
        kv_block_idx_bl = kv_idx // block_len
        is_in_bottom_left_block_causal_past = (q_idx >= half_seq_len) & (kv_idx < half_seq_len) & (q_block_idx_bl > kv_block_idx_bl)
    
        return is_in_top_left_causal | is_in_bottom_right_block | is_in_bottom_left_block_causal_past
    return create_block_mask(mask_mod, B=None, H=None, Q_LEN=seq_len, KV_LEN=seq_len)
\end{minted}
  \caption{FlexAttention implementation of the attention mask at training time (see \cref{fig:training} for a visual representation).}
  \label{app:code:train_attention_code}
\end{figure}

\clearpage

\begin{figure}[H]
  \centering
\begin{minted}{python}
from torch.nn.attention.flex_attention import BlockMask, create_block_mask

def create_attention_mask_inference(causal_point: int) -> BlockMask:
    # causal_point is the index of the first token in the prediction block  
    def mask_mod(b, h, q_idx, kv_idx):
        is_causal = (kv_idx <= q_idx)
        is_past_causal_point = (causal_point <= q_idx)
        return is_causal | is_past_causal_point
    return create_block_mask(mask_mod, B=None, H=None, Q_LEN=seq_len, KV_LEN=seq_len)
\end{minted}
  \caption{FlexAttention implementation of the attention mask at inference time. The attention is causal up until the point where the prediction block begins. The tokens in the prediction block attend to all other tokens in the sequence.}
  \label{app:code:inference_attention_code}
\end{figure}

\clearpage

\subsection{Roofline model}

\begin{figure}[H]
  \centering
\begin{minted}{python}
def gemm_roofline(m: int, n: int, k: int) -> tuple[float, float]:
    "Returns the runtime (in s) and flops of doing a `C=A@B` with A/B of shape `[m, k]/[k, n]` "
    total_IO = (
        m * k * BYTES + # read A
        n * k * BYTES + # read B
        m * n * BYTES   # write C
    )
    total_flops = 2 * m * n * k
    return max(total_IO / MEMORY_BANDWIDTH, total_flops / PEAK_FLOPS), total_flops

# Attention roofline analysis
def calculate_fused_attention_roofline(block_size: int, kv_length: int, batch_size: int) -> float:
    num_tokens = block_size * batch_size
    # Memory in Q: (batch_size, block_size,  NUM_HEADS, HEAD_DIM)
    bytes_Q = num_tokens * NUM_HEADS * HEAD_DIM * BYTES
    # Memory in K: (batch_size, kv_length, NUM_KV_HEADS, HEAD_DIM)
    bytes_K = batch_size * kv_length * NUM_KV_HEADS * HEAD_DIM * BYTES_PER_KV_ELEMENT
    # Flops P=QK^T
    _, flops_QKt = gemm_roofline(num_tokens, kv_length, NUM_HEADS * HEAD_DIM)    
    # Softmax memory transfer is ignored as it's not read or stored in memory, flops negliglbe
    bytes_softmax = 0
    # Memory in V: (batch_size, kv_length, NUM_KV_HEADS, HEAD_DIM)
    bytes_V = batch_size * kv_length * NUM_KV_HEADS * HEAD_DIM * BYTES_PER_KV_ELEMENT
    # Flops PV
    _, flops_PV = gemm_roofline(num_tokens, kv_length, NUM_HEADS * HEAD_DIM)    
    # Memory out PV: (batch_size, block_size, NUM_HEADS, HEAD_DIM)
    bytes_PV = num_tokens * NUM_HEADS * HEAD_DIM * BYTES

    total_memory = bytes_Q + bytes_K + bytes_V + bytes_PV + bytes_softmax
    total_flops = flops_QKt + flops_PV
    return max(total_memory / MEMORY_BANDWIDTH, total_flops / PEAK_FLOPS_ATTENTION)

# FFN roofline analysis
def calculate_linear_layers_roofline(block_size: int, batch_size: int) -> float:
    num_tokens = block_size * batch_size
    # FFN: Up-projection
    total_time = gemm_roofline(num_tokens, HIDDEN_DIM, HIDDEN_DIM * FFN_DIM)[0]
    # FFN: Down-projection
    total_time += gemm_roofline(num_tokens, HIDDEN_DIM * FFN_DIM, HIDDEN_DIM)[0]
    # Attention linear layers
    total_time += gemm_roofline(num_tokens, HIDDEN_DIM, NUM_HEADS * HEAD_DIM)[0] # Q-proj
    total_time += gemm_roofline(num_tokens, HIDDEN_DIM, NUM_HEADS * NUM_KV_HEADS)[0] # K-proj
    total_time += gemm_roofline(num_tokens, HIDDEN_DIM, NUM_HEADS * NUM_KV_HEADS)[0] # V-proj
    total_time += gemm_roofline(num_tokens, NUM_HEADS * HEAD_DIM, HIDDEN_DIM)[0] # out-proj
    return total_time
\end{minted}
  \caption{Transformer roofline analysis.}  \label{code:transformer_roofline}
\end{figure}

\section{Additional experiments}

\subsection{Sampling algorithm} \label{app:factor}

In all our experiments we used EB-Sampler in its simplest form with entropy error proxy (see \cref{e:eb_sampler}). In this section we ablate on two other sampling variants: (i) Factor parallel decoding \citep{wu2025fast}; and (ii) EB-Sampler with confidence error proxy \citep{ben2025accelerated}. For completeness, let us briefly describe the two. 

\begin{figure}[H]
  \centering
  \begin{tabular}{ccc}
     \footnotesize HumanEval  & \footnotesize MBPP &
     \footnotesize GSM8K \\
     \includegraphics[width=0.3\textwidth]{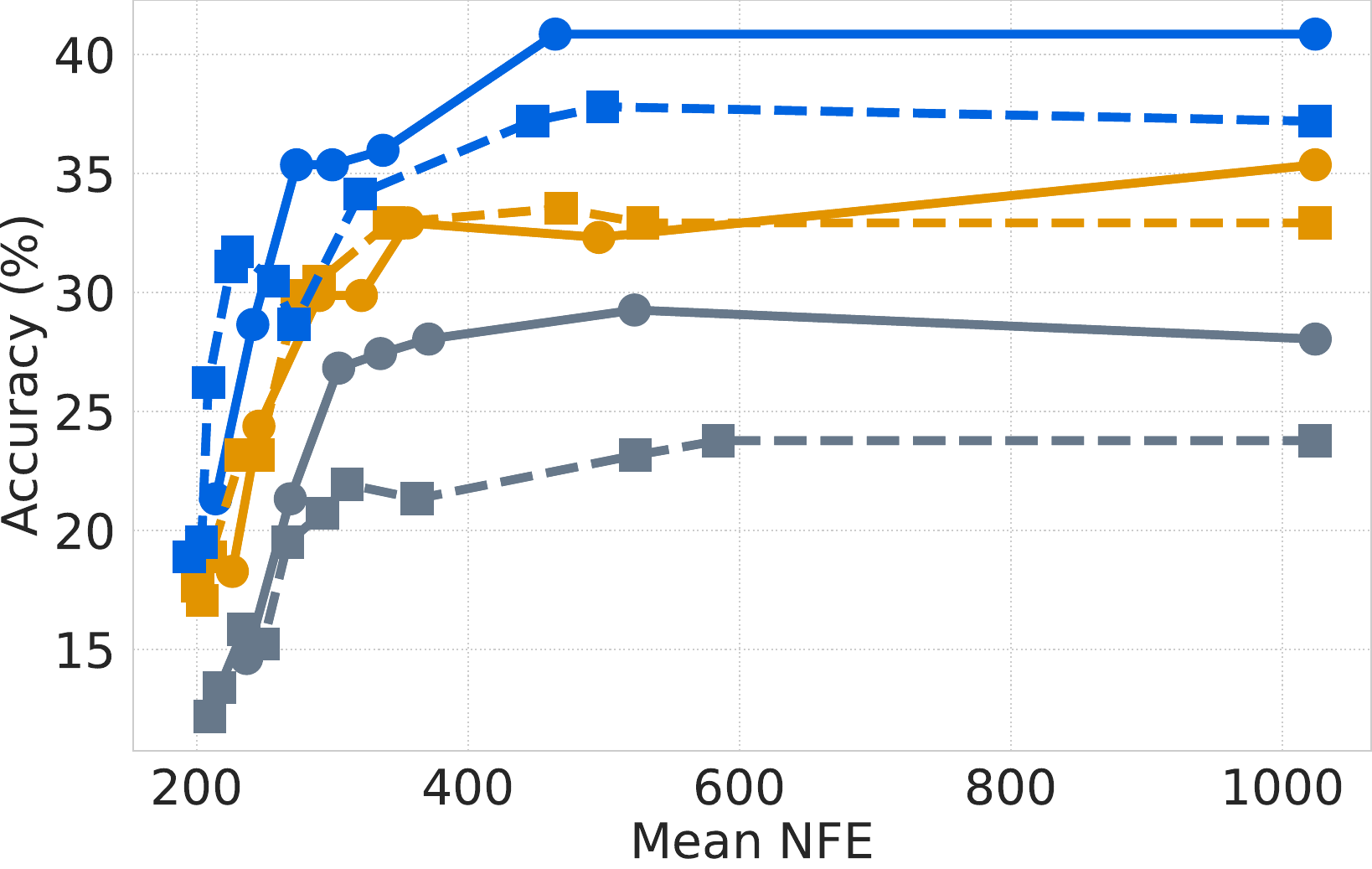}  & \includegraphics[width=0.3\textwidth]{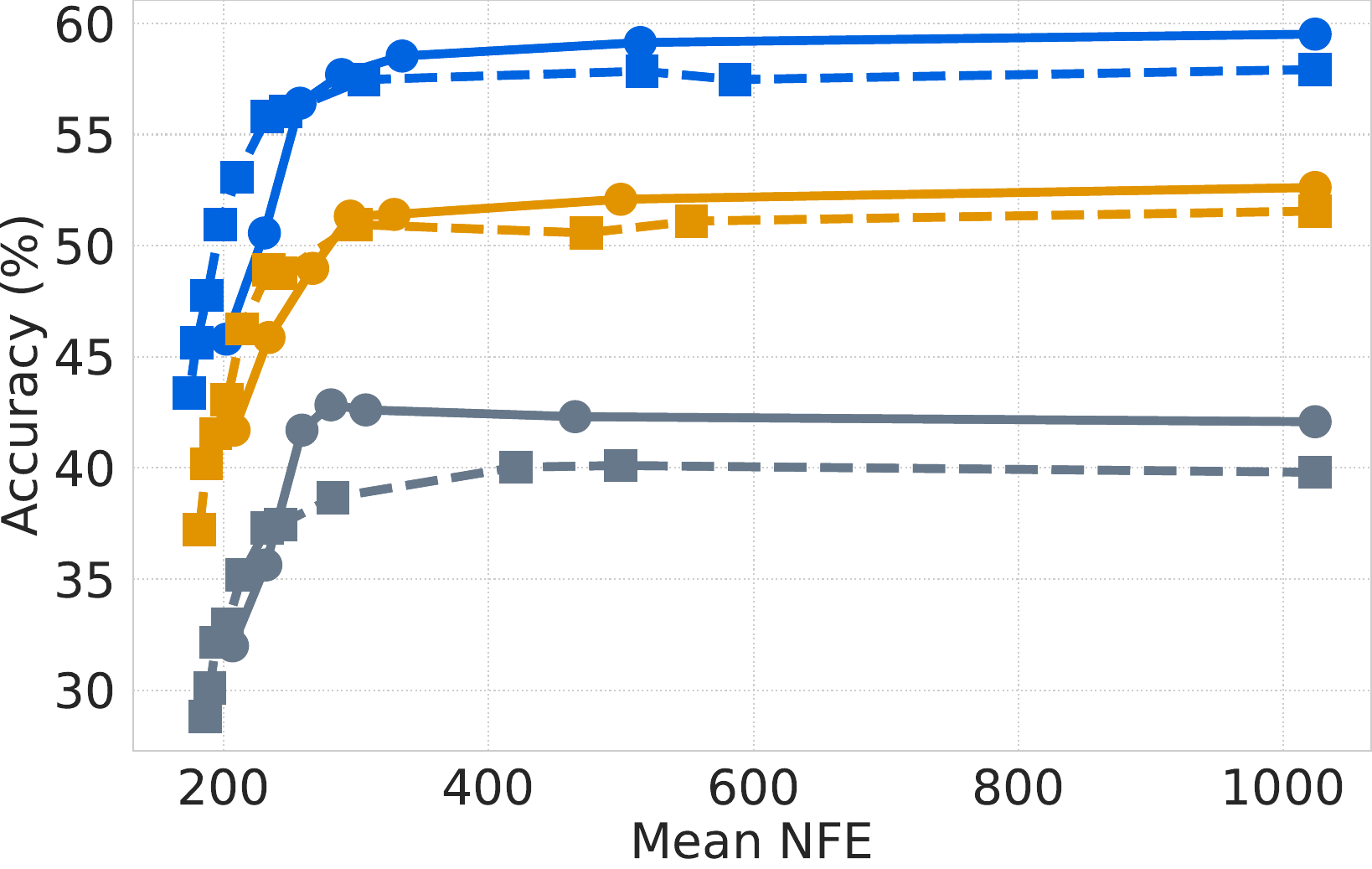} &     \includegraphics[width=0.3\textwidth]{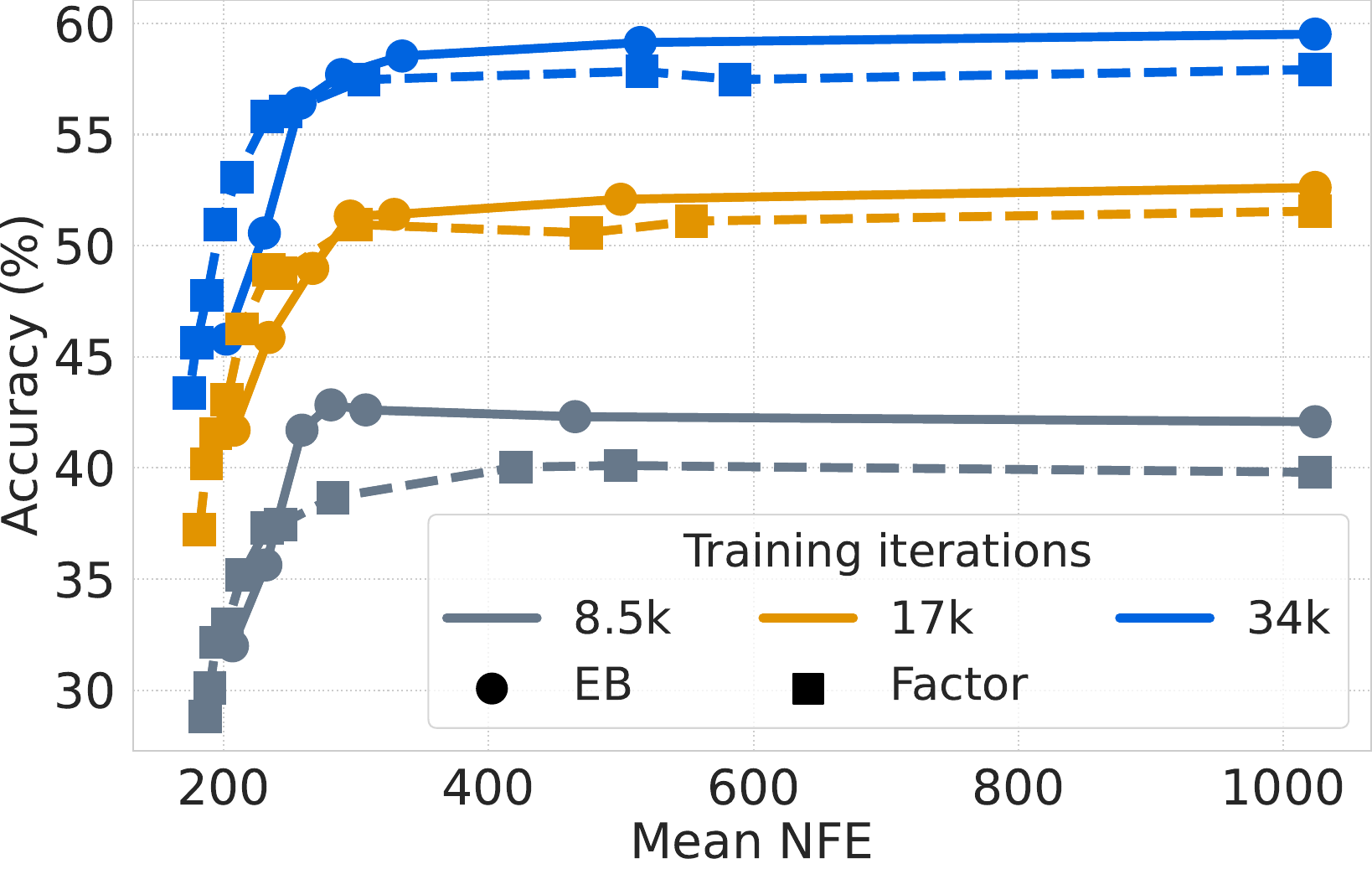}
  \end{tabular}
  \caption{3B model SFT training with varying number of training steps. Factor vs EB-Sampler with entropy error proxy.}
  \label{fig:3B_sft_ent}
  \vspace{-10pt}
\end{figure}

Recently, \cite{wu2025fast} proposed the Factor parallel decoding approach for sampling. With a similar motivation as the EB-Sampler, the Factor method proposes an adaptive parallel decoding algorithm that in high confidence regimes is equivalent to greedy decoding. At each step, given we already decoded the tokens in indices $\gJ \subset \gI$ and are left with masks $\gM=\gI \setminus \gJ$, compute the confidence for each masked index $c^i=\max_{x_i\in\gV} p(x_i|x_{<t},x_\gJ), i\in\gM$, sort the masked indices by confidence $i_1, i_2,\dots,i_{|\gM|}$ and unmask the $n$ first tokens. Where $n$ is the largest such that $(n+1)c^n<f$ and $f$ is a predefined threshold hyperparameter. In \cref{fig:3B_sft_ent} we compare the performance of the EB-Sampler with entropy error proxy to the factor method on the 3B SFT models from \cref{sec:ablation}. EB-Sampler with entropy error proxy compares favorably to the Factor method on HumanEval at all training budgets, where on MBPP and GSM8K both approaches perform similarly when models are trained for more steps.

Interestingly, the Factor method is more closely related to the EB-Sampler instance that uses confidence as an error proxy. In each step this instance sorts the masked tokens, $\gM$, by confidence, similarly to the Factor method, but chooses how many to unmask according to the entropy bound on the mutual information, finding the largest $s$ such that:
\begin{equation}\label{e:eb_sampler_general}
    \sum_{j=1}^{s} H(p(x_{i_j}|x_{<t},x_\gJ)) - \max_{j'\leq s} H(p(x_{i_{j'}}|x_{<t},x_\gJ)) \leq \gamma,
\end{equation}

\begin{figure}[H]
  \centering
  \begin{tabular}{ccc}
     \footnotesize HumanEval  & \footnotesize MBPP &
     \footnotesize GSM8K \\
     \includegraphics[width=0.3\textwidth]{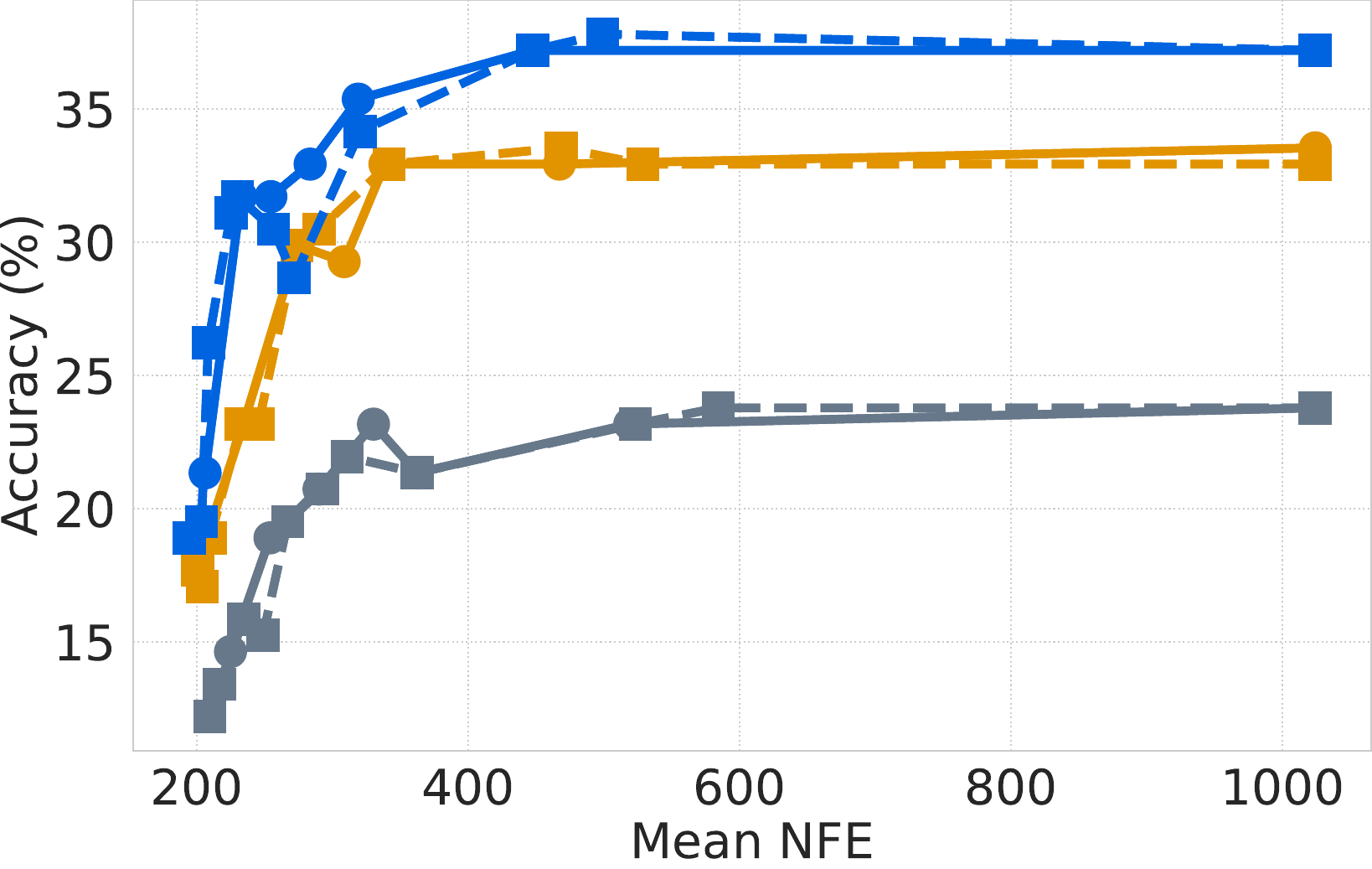}  & \includegraphics[width=0.3\textwidth]{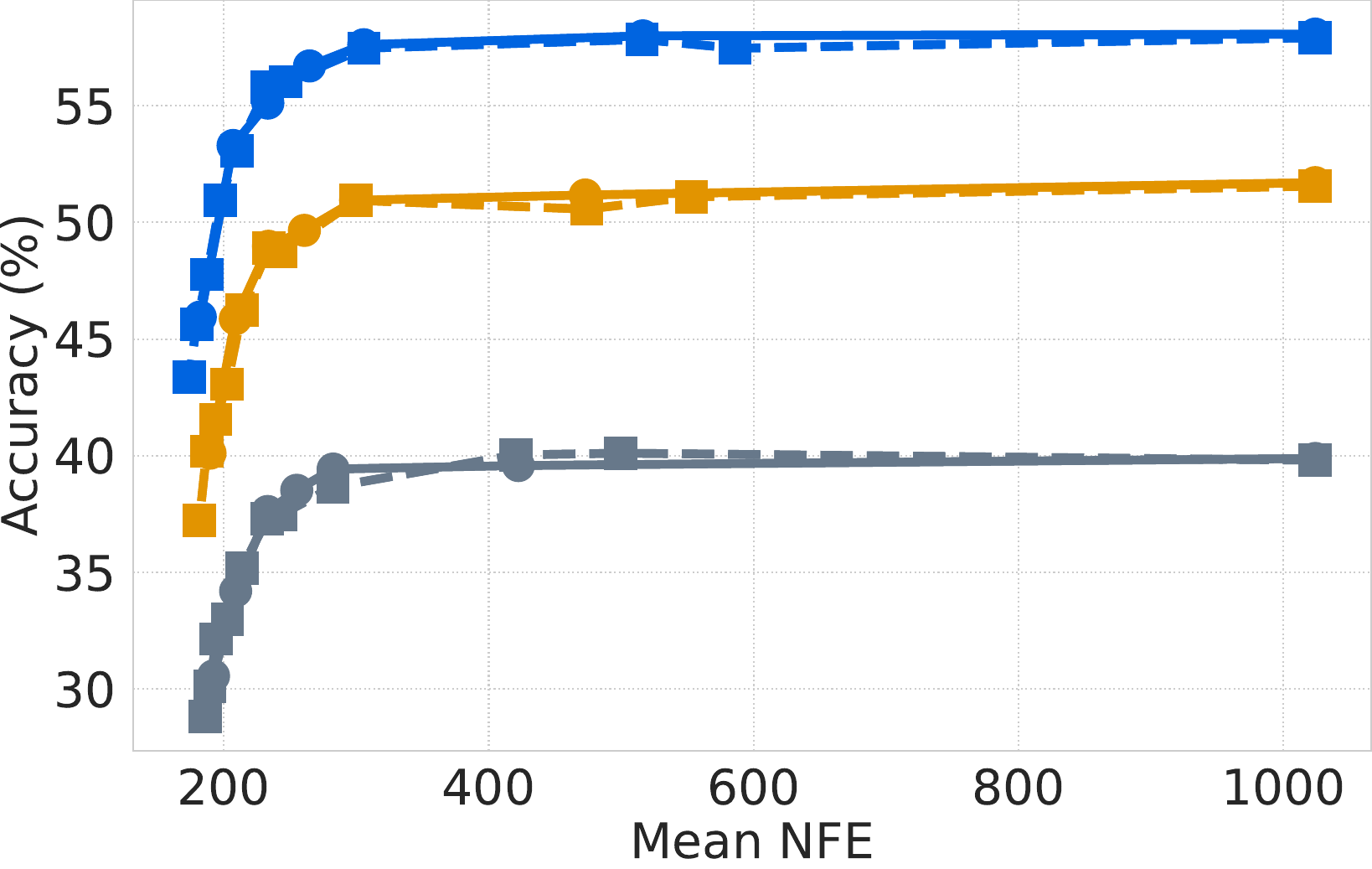} &     \includegraphics[width=0.3\textwidth]{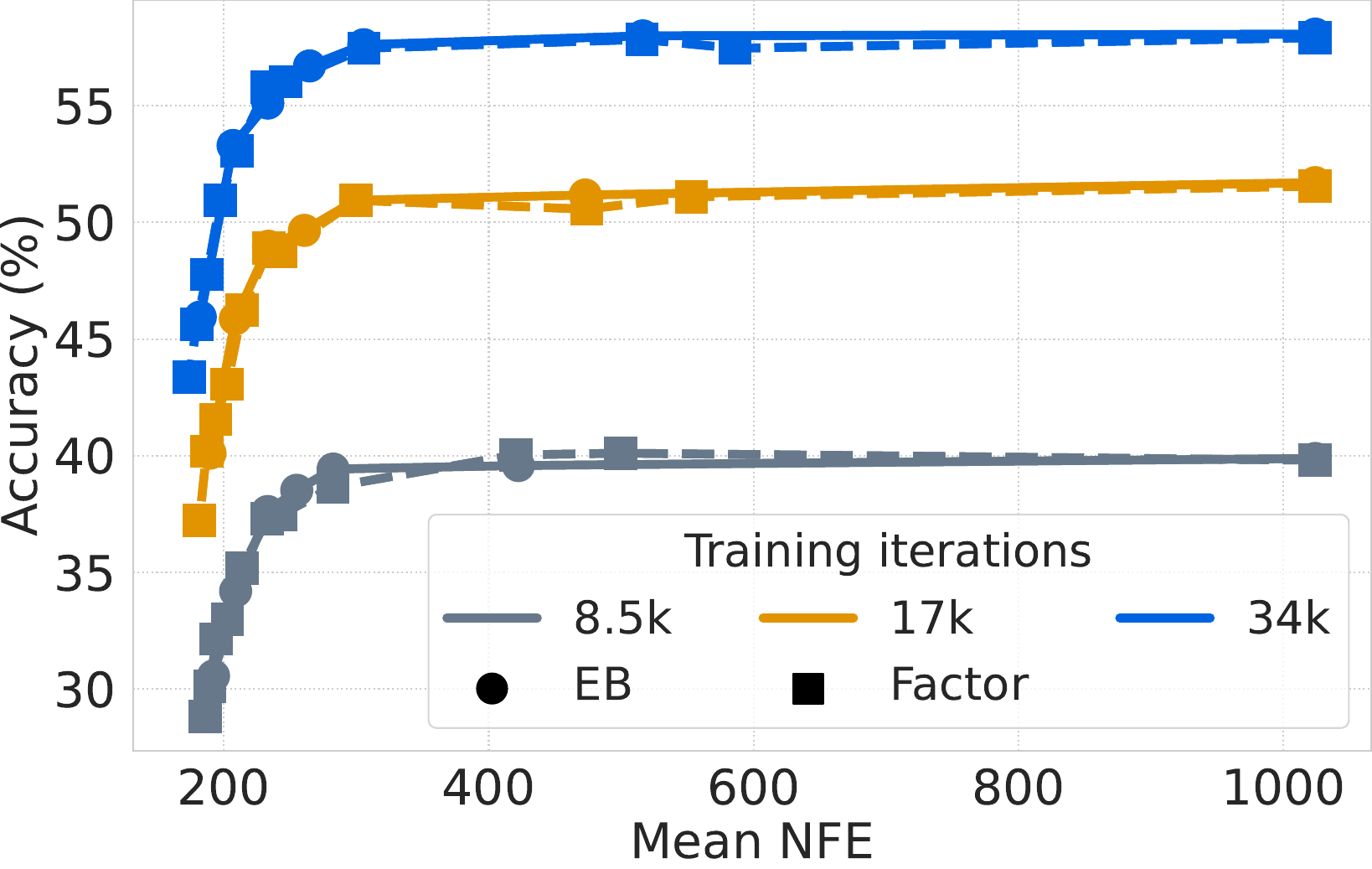}
  \end{tabular}
  \caption{3B model SFT training with varying number of training steps. Factor vs EB-Sampler with confidence error proxy.}
  \label{fig:3B_sft_conf}
  \vspace{-10pt}
\end{figure}

\Cref{fig:3B_sft_conf} shows the performance of the EB-Sampler with confidence error proxy compared to the Factor method at different $\gamma$ and $f$ values respectively, empirically demonstrating the similarity of the two approaches. Notably, \cref{fig:3B_sft_ent} and \cref{fig:3B_sft_conf} also show that entropy as an error proxy compares favorably for the tested models. The models used in this experiment are again the 3B SFT models from \cref{sec:ablation}.

\end{document}

%% file: math_commands.tex
\usepackage{amsmath}
\usepackage{enumerate} 
\usepackage{algorithm}
\usepackage{algpseudocode}
\usepackage{amsfonts}
\usepackage{amsthm}
\usepackage{newtxtt}
\usepackage{algorithm}
\usepackage{bm}
\usepackage{diagbox} 
\usepackage{colortbl}
\usepackage{nicematrix}
\usepackage{subcaption}
\usepackage{cellspace}
\usepackage{tcolorbox}
\usepackage[framemethod=tikz]{mdframed}
\usepackage{amssymb}
\usepackage{xspace}
\usepackage{wrapfig}
\usepackage{adjustbox}
\usepackage{tabularx}
\usepackage{booktabs}
\usepackage{mathtools}
\usepackage{tikz}
\usetikzlibrary{arrows.meta, bending} %
\usetikzlibrary{decorations.text}

\definecolor{pink}{RGB}{255, 204, 204}
\definecolor{bluel}{RGB}{230, 230, 255}
\definecolor{light}{RGB}{125, 125, 125}
\definecolor{pastelblue}{HTML}{cee4f6}
\definecolor{pastelgreen}{RGB}{204, 255, 204}
\definecolor{pastelpink}{HTML}{edc8e3}

\definecolor{darkerpink}{RGB}{220, 110, 110}
\definecolor{pastelgray}{RGB}{237, 237, 237}
\definecolor{faintred}{HTML}{FFDDDD}
\definecolor{lightred}{HTML}{FFBBBB}
\definecolor{darkred}{HTML}{FF9999}

\usepackage{silence}

\newcommand{\set}[1]{\left\{#1\right\}}

\newcommand{\Real}{\mathbb R}

\definecolor{mygray}{gray}{0.95}

\newcommand*{\eg}{{e.g.}\@\xspace}
\newcommand*{\ie}{{i.e.}\@\xspace}

\newcommand{\xr}[1]{\color{OliveGreen} {\tiny(#1x)}}
\newcommand{\mr}[1]{\color{BrickRed} {\tiny-#1\%}}
\newcommand{\pb}[1]{{\tiny+#1\%}}
\newcommand{\mb}[1]{{\tiny-#1\%}}
\newcommand{\er}[1]{\color{Gray} {\tiny +0.0\%}}

\theoremstyle{plain}

\theoremstyle{definition}

\theoremstyle{remark}

\makeatletter
\newtheorem*{rep@theorem}{\rep@title}
\newcommand{\newreptheorem}[2]{%
\newenvironment{rep#1}[1]{%
 \def\rep@title{\textbf{#2} \ref{##1}}%
 \begin{rep@theorem}}%
 {\end{rep@theorem}}}
\makeatother

\newreptheorem{theorem}{Theorem}
\newreptheorem{proposition}{Proposition}
\newreptheorem{lemma}{Lemma}
\newreptheorem{corollary}{Corollary}

\def\eqref#1{equation~\ref{#1}}

\def\1{\bm{1}}

\crefformat{equation}{#2equation~#1#3}
\Crefformat{equation}{#2Equation~#1#3}

\DeclareMathAlphabet{\mathsfit}{\encodingdefault}{\sfdefault}{m}{sl}
\SetMathAlphabet{\mathsfit}{bold}{\encodingdefault}{\sfdefault}{bx}{n}

\def\gD{{\mathcal{D}}}

\def\gI{{\mathcal{I}}}
\def\gJ{{\mathcal{J}}}

\def\gL{{\mathcal{L}}}
\def\gM{{\mathcal{M}}}

\def\gT{{\mathcal{T}}}

\def\gV{{\mathcal{V}}}

\newcolumntype{C}[1]{>{\Centering}m{#1}}

\newcolumntype{Z}[1]{>{\Left}m{#1}}

\algdef{SE}[REPEATN]{RepeatN}{End}[1]{\algorithmicrepeat\ #1 \textbf{times}}{\algorithmicend \textbf{ repeat}}

\usepackage[frozencache,cachedir=.]{minted}
\definecolor{LightGray}{gray}{0.97}
\setminted{
    frame=lines,
    framesep=2mm,
    baselinestretch=1.2,
    bgcolor=LightGray,
    fontsize=\footnotesize,
    breaklines,
    linenos
}

%% file: figures/algorithms.tex
\newcommand{\sample}{\textsc{sample\_block}}

\begin{center}
\resizebox{0.45\linewidth}{!}{%
\begin{minipage}[t]{0.65\textwidth}
    \begin{algorithm}[H] %
        \caption{Set Block Decoding training} \label{alg:bd_training}        
        \begin{algorithmic}[1]
            \Require Init model params $\theta$, data $\gD$.
    \For{data sequence $x \in \mathcal{D}$}            
            \State Random noising probability $\tau \sim U(0,1)$
            \State Create a masked sequence $\hat{x}$, see \eqref{e:masked_sequence}
            \State Compute loss $\gL(x,\hat{x};\theta)$, see \eqref{e:loss}
            \State Compute gradients and update parameters $\theta$
    \EndFor
    \State \textbf{return} $\theta$
        \end{algorithmic}
    \end{algorithm}
\end{minipage}
}
\hspace{20pt}
\resizebox{0.45\linewidth}{!}{%
\begin{minipage}[t]{0.65\textwidth}
    \begin{algorithm}[H] %
        \caption{Set Block Decoding inference}
        \label{alg:bd_inference}
        \begin{algorithmic}[1]
            \Require Model $f_\theta$, prompt $x_{<0}$, block size $k$, length $L$
    \State $x \leftarrow x_{<0}$
    \State Prefill $f_\theta(x;)$    
    \For{$t=0,k,2k,\ldots,L-k$}
        \State $x_{<t+k} \leftarrow \sample(x_{<t})$
    \EndFor    
    \State \textbf{return} $x_{<L}$
    \vspace{10pt}
        \end{algorithmic}
    \end{algorithm}
\end{minipage} }
\end{center}

%% file: figures/model_attention.tex
\resizebox{0.3\textwidth}{!}{
\begin{tblr}{
  width = 1.8\linewidth, 
  colspec = {| X[c,m] | X[c,m] X[c,m] X[c,m] X[c,m] X[c,m]  | X[c,m]  X[c,m] X[c,m] X[c,m] |},
  rowspec = {Q[m]Q[m]Q[m]},
  rows = {ht=0.5cm}, %
  row{1} = {gray!10},
  column{1} = {gray!10},
  cell{2}{2} = {bg=pastelblue!70},
  cell{3}{2-3} = {bg=pastelblue!70},
  cell{4}{2-4} = {bg=pastelblue!70},
  cell{5}{2-5} = {bg=pastelblue!70},
  cell{6}{2-6} = {bg=pastelblue!70},
  cell{7}{2-10} = {bg=pastelblue!70},
  cell{8}{2-10} = {bg=pastelblue!70},
  cell{9}{2-10} = {bg=pastelblue!70},
  cell{10}{2-10} = {bg=pastelblue!70},
}
\hline %
 & $x_1$ & $x_2$ & $x_3$ & $\cdots$ & $x_{t-1}$ & $\hat{x}_{t}$ & $\hat{x}_{t+1}$ & $\hat{x}_{t+2}$ & $\hat{x}_{t+3}$  \\
\hline 
$x_1$ &  &  & & & & & & &  \\
$x_2$ &  &  & & & & & & &  \\
$x_3$ &  &  & & & & & & &  \\
$\vdots$ &  &  & & & & & & &  \\
$x_{t-1}$ &  &  & & & & & & &  \\
\hline %
$\hat{x}_{t}$ &  &  & & & & & & &  \\
$\hat{x}_{t+1}$ &  &  & & & & & & &  \\
$\hat{x}_{t+2}$ &  &  & & & & & & &  \\
$\hat{x}_{t+3}$ &  &  & & & & & & &  \\
\hline %
\end{tblr}
}

%% file: figures/sample_block.tex
\begin{wrapfigure}{r}{0.45\textwidth}
    \vspace{-5pt} %
    \resizebox{\linewidth}{!}{%
\begin{minipage}{0.65\textwidth}
    \begin{algorithm}[H]
        \caption{Sample block}
        \label{alg:block_sample}
        \begin{algorithmic}[1]
            \Require Model $f_\theta$, input sequence $x_{<t}$, block size $k$ 
            \State $\hat{x}_{t:t+k-1}=(\text{m},\ldots,\text{m})$ \Comment{init with $k$ mask tokens} 
            \While{$\hat{x}_{t:t+k-1}$ contains masks}
                \If{first iteration and $t\geq k$}                    
                    \State $\hat{z}_{t:t+k-1} \leftarrow f_\theta(x_{<t-k},\overbrace{\textcolor{darkerpink}{ x_{t-k:t-1}}}^{\tiny \text{KV-cache}};\hat{x}_{t:t+k-1})$ 
                \Else
                    \State $\hat{z}_{t:t+k-1} \leftarrow f_\theta(x_{<t};\hat{x}_{t:t+k-1})$
                \EndIf
                \State Compute probabilities from $\hat{z}_{t:t+k-1}$
                \State Unmask tokens in $\hat{x}_{t:t+k-1}$ according to \ref{e:eb_sampler}
            \EndWhile       
            \State \Return $(x_{<t},\hat{x}_{t:t+k-1})$ 
        \end{algorithmic}
    \end{algorithm}
    \end{minipage}
    }
\end{wrapfigure}

%% file: figures/roofline/tables/slowdown_table_batch_1.tex
\begin{tabular}{lccccccc}
\toprule
\multirow{2}{*}{KV Cache Length} & \multicolumn{7}{c}{Block Size} \\
\cmidrule(r){2-8}
 & 1 & 2 & 4 & 8 & 16 & 32 & 64 \\
\midrule
$2^{4}$ (16) & 1.000 & 1.000 & 1.001 & 1.002 & 1.004 & 1.008 & 1.015 \\
$2^{8}$ (256) & 1.000 & 1.000 & 1.001 & 1.002 & 1.004 & 1.008 & 1.015 \\
$2^{10}$ (1,024) & 1.000 & 1.000 & 1.001 & 1.002 & 1.004 & 1.008 & 1.015 \\
$2^{12}$ (4,096) & 1.000 & 1.000 & 1.001 & 1.002 & 1.004 & 1.008 & 1.015 \\
$2^{14}$ (16,384) & 1.000 & 1.000 & 1.001 & 1.002 & 1.004 & 1.008 & 1.015 \\
$2^{16}$ (65,536) & 1.000 & 1.000 & 1.001 & 1.002 & 1.004 & 1.008 & 1.016 \\
$2^{20}$ (1,048,576) & 1.000 & 1.000 & 1.001 & 1.002 & 1.004 & 1.008 & 1.019 \\
$2^{22}$ (4,194,304) & 1.000 & 1.000 & 1.001 & 1.002 & 1.004 & 1.007 & 1.029 \\
\bottomrule
\end{tabular}

%% file: figures/roofline/tables/slowdown_table_batch_4.tex
\begin{tabular}{lccccccc}
\toprule
\multirow{2}{*}{KV Cache Length} & \multicolumn{7}{c}{Block Size} \\
\cmidrule(r){2-8}
 & 1 & 2 & 4 & 8 & 16 & 32 & 64 \\
\midrule
$2^{4}$ (16) & 1.000 & 1.001 & 1.003 & 1.007 & 1.015 & 1.030 & 1.061 \\
$2^{8}$ (256) & 1.000 & 1.001 & 1.003 & 1.007 & 1.015 & 1.030 & 1.061 \\
$2^{10}$ (1,024) & 1.000 & 1.001 & 1.003 & 1.007 & 1.015 & 1.030 & 1.061 \\
$2^{12}$ (4,096) & 1.000 & 1.001 & 1.003 & 1.007 & 1.015 & 1.030 & 1.062 \\
$2^{14}$ (16,384) & 1.000 & 1.001 & 1.003 & 1.007 & 1.015 & 1.030 & 1.062 \\
$2^{16}$ (65,536) & 1.000 & 1.001 & 1.003 & 1.007 & 1.015 & 1.030 & 1.062 \\
$2^{20}$ (1,048,576) & 1.000 & 1.001 & 1.003 & 1.007 & 1.014 & 1.030 & 1.074 \\
$2^{22}$ (4,194,304) & 1.000 & 1.001 & 1.003 & 1.006 & 1.014 & 1.028 &  1.111 \\
\bottomrule
\end{tabular}

%% file: figures/roofline/tables/slowdown_table_batch_8.tex
\begin{tabular}{lccccccc}
\toprule
\multirow{2}{*}{KV Cache Length} & \multicolumn{7}{c}{Block Size} \\
\cmidrule(r){2-8}
 & 1 & 2 & 4 & 8 & 16 & 32 & 64 \\
\midrule
$2^{4}$ (16) & 1.000 & 1.002 & 1.006 & 1.014 & 1.029 & 1.060 & \cellcolor{faintred} 1.903 \\
$2^{8}$ (256) & 1.000 & 1.002 & 1.006 & 1.014 & 1.029 & 1.060 & \cellcolor{faintred}1.903 \\
$2^{10}$ (1,024) & 1.000 & 1.002 & 1.006 & 1.014 & 1.029 & 1.060 & \cellcolor{faintred}1.903 \\
$2^{12}$ (4,096) & 1.000 & 1.002 & 1.006 & 1.014 & 1.029 & 1.060 & \cellcolor{faintred}1.903 \\
$2^{14}$ (16,384) & 1.000 & 1.002 & 1.006 & 1.014 & 1.029 & 1.060 & \cellcolor{faintred}1.903 \\
$2^{16}$ (65,536) & 1.000 & 1.002 & 1.006 & 1.014 & 1.029 & 1.060 & \cellcolor{faintred}1.903 \\
$2^{20}$ (1,048,576) & 1.000 & 1.002 & 1.006 & 1.013 & 1.028 & 1.059 & \cellcolor{faintred}1.903 \\
$2^{22}$ (4,194,304) & 1.000 & 1.002 & 1.005 & 1.012 & 1.026 & 1.054 & \cellcolor{faintred}1.904 \\
\bottomrule
\end{tabular}

%% file: figures/roofline/tables/slowdown_table_batch_16.tex
\begin{tabular}{lccccccc}
\toprule
\multirow{2}{*}{KV Cache Length} & \multicolumn{7}{c}{Block Size} \\
\cmidrule(r){2-8}
 & 1 & 2 & 4 & 8 & 16 & 32 & 64 \\
\midrule
$2^{4}$ (16) & 1.000 & 1.004 & 1.012 & 1.027 & 1.058 & \cellcolor{faintred} 1.899 & \cellcolor{lightred} 3.799 \\
$2^{8}$ (256) & 1.000 & 1.004 & 1.012 & 1.027 & 1.058 & \cellcolor{faintred}1.899 &\cellcolor{lightred}3.799 \\
$2^{10}$ (1,024) & 1.000 & 1.004 & 1.012 & 1.027 & 1.058 & \cellcolor{faintred}1.899 & \cellcolor{lightred}3.799 \\
$2^{12}$ (4,096) & 1.000 & 1.004 & 1.012 & 1.027 & 1.058 & \cellcolor{faintred}1.899 &\cellcolor{lightred} 3.798 \\
$2^{14}$ (16,384) & 1.000 & 1.004 & 1.012 & 1.027 & 1.058 & \cellcolor{faintred}1.899 &\cellcolor{lightred} 3.797 \\
$2^{16}$ (65,536) & 1.000 & 1.004 & 1.012 & 1.027 & 1.058 & \cellcolor{faintred}1.896 & \cellcolor{lightred}3.792 \\
$2^{20}$ (1,048,576) & 1.000 & 1.004 & 1.011 & 1.026 & 1.055 & \cellcolor{faintred}1.847 &\cellcolor{lightred} 3.688 \\
$2^{22}$ (4,194,304) & 1.000 & 1.003 & 1.009 & 1.022 & 1.047 & \cellcolor{faintred}1.720 &\cellcolor{lightred} 3.422 \\
\bottomrule
\end{tabular}

%% file: figures/roofline/tables/speedup_table_batch_1_nfe_speedup_2.tex
\begin{tabular}{lcccc}
\toprule
\multirow{2}{*}{KV Cache Length} & \multicolumn{4}{c}{Block Size} \\
\cmidrule(r){2-5}
 & 8 & 16 & 32 & 64 \\
\midrule
$2^{4}$ (16) & 1.996 & 1.992 & 1.984 & 1.969 \\
$2^{8}$ (256) & 1.996 & 1.992 & 1.984 & 1.969 \\
$2^{10}$ (1,024) & 1.996 & 1.992 & 1.984 & 1.969 \\
$2^{12}$ (4,096) & 1.996 & 1.992 & 1.984 & 1.969 \\
$2^{14}$ (16,384) & 1.996 & 1.992 & 1.984 & 1.969 \\
$2^{16}$ (65,536) & 1.996 & 1.992 & 1.984 & 1.968 \\
$2^{20}$ (1,048,576) & 1.996 & 1.992 & 1.984 & 1.962 \\
$2^{22}$ (4,194,304) & 1.996 & 1.992 & 1.983 & 1.941 \\
\bottomrule
\end{tabular}

%% file: figures/roofline/tables/speedup_table_batch_4_nfe_speedup_2.tex
\begin{tabular}{lcccc}
\toprule
\multirow{2}{*}{KV Cache Length} & \multicolumn{4}{c}{Block Size} \\
\cmidrule(r){2-5}
 & 8 & 16 & 32 & 64 \\
\midrule
$2^{4}$ (16) & 1.983 & 1.967 & 1.938 & \cellcolor{faintred} 1.839 \\
$2^{8}$ (256) & 1.983 & 1.967 & 1.938 & \cellcolor{faintred}1.839 \\
$2^{10}$ (1,024) & 1.983 & 1.967 & 1.938 & \cellcolor{faintred}1.838 \\
$2^{12}$ (4,096) & 1.983 & 1.967 & 1.938 & \cellcolor{faintred}1.838 \\
$2^{14}$ (16,384) & 1.983 & 1.967 & 1.938 &\cellcolor{faintred} 1.838 \\
$2^{16}$ (65,536) & 1.983 & 1.967 & 1.938 & \cellcolor{faintred}1.837 \\
$2^{20}$ (1,048,576) & 1.983 & 1.968 & 1.937 & \cellcolor{faintred}1.816 \\
$2^{22}$ (4,194,304) & 1.984 & 1.969 & 1.935 &\cellcolor{faintred} 1.755 \\
\bottomrule
\end{tabular}

%% file: figures/roofline/tables/speedup_table_batch_8_nfe_speedup_2.tex
\begin{tabular}{lcccc}
\toprule
\multirow{2}{*}{KV Cache Length} & \multicolumn{4}{c}{Block Size} \\
\cmidrule(r){2-5}
 & 8 & 16 & 32 & 64 \\
\midrule
$2^{4}$ (16) & 1.966 & 1.936 & \cellcolor{faintred}1.797 & \cellcolor{darkred}1.019 \\
$2^{8}$ (256) & 1.966 & 1.936 & \cellcolor{faintred}1.797 &\cellcolor{darkred} 1.019 \\
$2^{10}$ (1,024) & 1.966 & 1.936 &\cellcolor{faintred} 1.797 & \cellcolor{darkred}1.019 \\
$2^{12}$ (4,096) & 1.966 & 1.936 & \cellcolor{faintred}1.797 & \cellcolor{darkred}1.019 \\
$2^{14}$ (16,384) & 1.966 & 1.936 &\cellcolor{faintred} 1.797 & \cellcolor{darkred}1.019 \\
$2^{16}$ (65,536) & 1.966 & 1.936 & \cellcolor{faintred}1.797 & \cellcolor{darkred}1.019 \\
$2^{20}$ (1,048,576) & 1.967 & 1.938 &\cellcolor{faintred} 1.800 &\cellcolor{darkred} 1.019 \\
$2^{22}$ (4,194,304) & 1.969 & 1.943 & \cellcolor{faintred}1.807 &\cellcolor{darkred} 1.019 \\
\bottomrule
\end{tabular}

%% file: figures/roofline/tables/speedup_table_batch_1_nfe_speedup_4.tex
\begin{tabular}{lcccc}
\toprule
\multirow{2}{*}{KV Cache Length} & \multicolumn{4}{c}{Block Size} \\
\cmidrule(r){2-5}
 & 8 & 16 & 32 & 64 \\
\midrule
$2^{4}$ (16) & 3.989 & 3.982 & 3.966 & 3.936 \\
$2^{8}$ (256) & 3.989 & 3.982 & 3.966 & 3.936 \\
$2^{10}$ (1,024) & 3.989 & 3.982 & 3.966 & 3.936 \\
$2^{12}$ (4,096) & 3.989 & 3.982 & 3.966 & 3.936 \\
$2^{14}$ (16,384) & 3.989 & 3.982 & 3.966 & 3.935 \\
$2^{16}$ (65,536) & 3.989 & 3.982 & 3.966 & 3.935 \\
$2^{20}$ (1,048,576) & 3.989 & 3.982 & 3.965 & 3.920 \\
$2^{22}$ (4,194,304) & 3.989 & 3.982 & 3.960 & 3.876 \\
\bottomrule
\end{tabular}

%% file: figures/roofline/tables/speedup_table_batch_4_nfe_speedup_4.tex
\begin{tabular}{lcccc}
\toprule
\multirow{2}{*}{KV Cache Length} & \multicolumn{4}{c}{Block Size} \\
\cmidrule(r){2-5}
 & 8 & 16 & 32 & 64 \\
\midrule
$2^{4}$ (16) & 3.958 & 3.927 & \cellcolor{faintred} 3.868 &\cellcolor{lightred} 3.590 \\
$2^{8}$ (256) & 3.958 & 3.927 &\cellcolor{faintred} 3.868 & \cellcolor{lightred}3.590 \\
$2^{10}$ (1,024) & 3.958 & 3.927 &\cellcolor{faintred} 3.868 & \cellcolor{lightred}3.590 \\
$2^{12}$ (4,096) & 3.958 & 3.927 &\cellcolor{faintred} 3.868 & \cellcolor{lightred}3.590 \\
$2^{14}$ (16,384) & 3.958 & 3.927 &\cellcolor{faintred} 3.868 & \cellcolor{lightred}3.589 \\
$2^{16}$ (65,536) & 3.958 & 3.927 &\cellcolor{faintred} 3.868 & \cellcolor{lightred}3.587 \\
$2^{20}$ (1,048,576) & 3.958 & 3.928 &\cellcolor{faintred} 3.863 & \cellcolor{lightred}3.545 \\
$2^{22}$ (4,194,304) & 3.960 & 3.931 &\cellcolor{faintred} 3.851 & \cellcolor{lightred}3.425 \\
\bottomrule
\end{tabular}

%% file: figures/roofline/tables/speedup_table_batch_8_nfe_speedup_4.tex
\begin{tabular}{lcccc}
\toprule
\multirow{2}{*}{KV Cache Length} & \multicolumn{4}{c}{Block Size} \\
\cmidrule(r){2-5}
 & 8 & 16 & 32 & 64 \\
\midrule
$2^{4}$ (16) & 3.916 & \cellcolor{faintred} 3.857 & \cellcolor{lightred}3.431 & \cellcolor{darkred}1.978 \\
$2^{8}$ (256) & 3.916 & \cellcolor{faintred}3.857 & \cellcolor{lightred}3.431 & \cellcolor{darkred}1.978 \\
$2^{10}$ (1,024) & 3.916 & \cellcolor{faintred}3.857 & \cellcolor{lightred}3.431 & \cellcolor{darkred}1.978 \\
$2^{12}$ (4,096) & 3.916 & \cellcolor{faintred}3.857 & \cellcolor{lightred}3.431 &\cellcolor{darkred} 1.978 \\
$2^{14}$ (16,384) & 3.916 & \cellcolor{faintred}3.857 &\cellcolor{lightred} 3.431 & \cellcolor{darkred}1.978 \\
$2^{16}$ (65,536) & 3.916 & \cellcolor{faintred}3.857 & \cellcolor{lightred}3.431 &\cellcolor{darkred} 1.978 \\
$2^{20}$ (1,048,576) & 3.918 & \cellcolor{faintred}3.861 & \cellcolor{lightred}3.436 &\cellcolor{darkred} 1.978 \\
$2^{22}$ (4,194,304) & 3.925 & \cellcolor{faintred}3.872 & \cellcolor{lightred}3.448 &\cellcolor{darkred} 1.978 \\
\bottomrule
\end{tabular}

%% file: figures/roofline/tables/speedup_table_batch_1_nfe_speedup_8.tex
\begin{tabular}{lcccc}
\toprule
\multirow{2}{*}{KV Cache Length} & \multicolumn{4}{c}{Block Size} \\
\cmidrule(r){2-5}
 & 8 & 16 & 32 & 64 \\
\midrule
$2^{4}$ (16) & 7.971 & 7.955 & 7.925 & \cellcolor{faintred} 7.864 \\
$2^{8}$ (256) & 7.971 & 7.955 & 7.925 & \cellcolor{faintred}7.864 \\
$2^{10}$ (1,024) & 7.971 & 7.955 & 7.925 & \cellcolor{faintred}7.864 \\
$2^{12}$ (4,096) & 7.971 & 7.955 & 7.925 &\cellcolor{faintred} 7.864 \\
$2^{14}$ (16,384) & 7.971 & 7.955 & 7.924 & \cellcolor{faintred}7.863 \\
$2^{16}$ (65,536) & 7.971 & 7.955 & 7.924 & \cellcolor{faintred}7.862 \\
$2^{20}$ (1,048,576) & 7.971 & 7.956 & 7.918 & \cellcolor{faintred}7.830 \\
$2^{22}$ (4,194,304) & 7.971 & 7.956 & 7.898 & \cellcolor{faintred}7.732 \\
\bottomrule
\end{tabular}

%% file: figures/roofline/tables/speedup_table_batch_4_nfe_speedup_8.tex
\begin{tabular}{lcccc}
\toprule
\multirow{2}{*}{KV Cache Length} & \multicolumn{4}{c}{Block Size} \\
\cmidrule(r){2-5}
 & 8 & 16 & 32 & 64 \\
\midrule
$2^{4}$ (16) & \cellcolor{faintred}7.885 & \cellcolor{faintred}7.824 & \cellcolor{faintred}7.707 &\cellcolor{lightred} 6.856 \\
$2^{8}$ (256) & \cellcolor{faintred}7.885 & \cellcolor{faintred}7.824 & \cellcolor{faintred}7.707 & \cellcolor{lightred}6.856 \\
$2^{10}$ (1,024) & \cellcolor{faintred}7.885 & \cellcolor{faintred}7.824 & \cellcolor{faintred}7.707 & \cellcolor{lightred}6.856 \\
$2^{12}$ (4,096) & \cellcolor{faintred}7.885 & \cellcolor{faintred}7.824 & \cellcolor{faintred}7.707 & \cellcolor{lightred}6.855 \\
$2^{14}$ (16,384) & \cellcolor{faintred}7.885 &\cellcolor{faintred} 7.824 & \cellcolor{faintred}7.706 &\cellcolor{lightred} 6.854 \\
$2^{16}$ (65,536) & \cellcolor{faintred}7.885 & \cellcolor{faintred}7.825 & \cellcolor{faintred}7.705 & \cellcolor{lightred}6.850 \\
$2^{20}$ (1,048,576) & \cellcolor{faintred}7.886 & \cellcolor{faintred}7.827 & \cellcolor{faintred}7.685 & \cellcolor{lightred}6.768 \\
$2^{22}$ (4,194,304) & \cellcolor{faintred}7.891 & \cellcolor{faintred}7.834 & \cellcolor{faintred}7.625 & \cellcolor{lightred}6.534 \\
\bottomrule
\end{tabular}

%% file: figures/roofline/tables/speedup_table_batch_8_nfe_speedup_8.tex
\begin{tabular}{lcccc}
\toprule
\multirow{2}{*}{KV Cache Length} & \multicolumn{4}{c}{Block Size} \\
\cmidrule(r){2-5}
 & 8 & 16 & 32 & 64 \\
\midrule
$2^{4}$ (16) & \cellcolor{faintred}7.773 & \cellcolor{faintred}7.657 & \cellcolor{lightred}6.294 & \cellcolor{darkred}3.736 \\
$2^{8}$ (256) & \cellcolor{faintred}7.773 & \cellcolor{faintred}7.657 & \cellcolor{lightred}6.294 & \cellcolor{darkred}3.736 \\
$2^{10}$ (1,024) & \cellcolor{faintred}7.773 & \cellcolor{faintred}7.657 & \cellcolor{lightred}6.294 & \cellcolor{darkred}3.736 \\
$2^{12}$ (4,096) & \cellcolor{faintred}7.773 & \cellcolor{faintred}7.657 &\cellcolor{lightred} 6.294 & \cellcolor{darkred}3.736 \\
$2^{14}$ (16,384) & \cellcolor{faintred}7.773 & \cellcolor{faintred}7.657 & \cellcolor{lightred}6.294 & \cellcolor{darkred}3.736 \\
$2^{16}$ (65,536) & \cellcolor{faintred}7.773 & \cellcolor{faintred}7.657 & \cellcolor{lightred}6.294 & \cellcolor{darkred}3.736 \\
$2^{20}$ (1,048,576) & \cellcolor{faintred}7.779 &\cellcolor{faintred} 7.667 & \cellcolor{lightred}6.300 & \cellcolor{darkred}3.736 \\
$2^{22}$ (4,194,304) & \cellcolor{faintred}7.797 & \cellcolor{faintred}7.693 & \cellcolor{lightred}6.318 & \cellcolor{darkred}3.736 \\
\bottomrule
\end{tabular}

%% file: figures/training_diagram.tex
\begin{figure}[h!]
\centering

\resizebox{0.7\linewidth}{!}{%
\begin{tblr}{
  width = 1.0\linewidth, 
  colspec = {| Q[20pt, c, m] | Q[20pt, c, m] | X[c,m] X[c,m] X[c,m] X[c,m]  | X[c,m]  X[c,m] X[c,m] X[c,m] | X[c,m]  X[c,m] X[c,m] X[c,m] | X[c,m]  X[c,m] X[c,m] X[c,m] |},
  rowspec = {Q[m]Q[m]Q[m]},
  rows = {ht=0.5cm}, %
  row{1} = {gray!10},
  column{1-2} = {gray!10},
  cell{2}{3} = {bg=pastelblue!70},
  cell{3}{3-4} = {bg=pastelblue!70},
  cell{4}{3-5} = {bg=pastelblue!70},
  cell{5}{3-6} = {bg=pastelblue!70},
  cell{6}{7-10} = {bg=pastelblue!70},
  cell{7}{7-10} = {bg=pastelblue!70},
  cell{8}{7-10} = {bg=pastelblue!70},
  cell{9}{7-10} = {bg=pastelblue!70},
  cell{10}{3-6, 11} = {bg=pastelblue!70},  
  cell{11}{3-6, 11-12} = {bg=pastelblue!70},  
  cell{12}{3-6, 11-13} = {bg=pastelblue!70},  
  cell{13}{3-6, 11-14} = {bg=pastelblue!70},  
  cell{14}{3-6, 15-18} = {bg=pastelblue!70},  
  cell{15}{3-6, 15-18} = {bg=pastelblue!70},  
  cell{16}{3-6, 15-18} = {bg=pastelblue!70},  
  cell{17}{3-6, 15-18} = {bg=pastelblue!70},  
}
\hline %
\scriptsize{Input} & \scriptsize{Target} & $x_1$ & $x_2$ & $x_3$ & $x_4$ & $x_1$ & \text{m} & $x_3$ & \text{m} & $x_5$ & $x_6$ & $x_7$ & $x_8$ & \text{m} & \text{m} & $x_7$ & \text{m} \\
\hline 
$x_1$ & $x_2$ &  & & & & & & &  \\
$x_2$ & $x_3$ &  & & & & & & &  \\
$x_3$ & $x_4$ &  & & & & & & &  \\
$x_4$ & $x_5$  &  & & & & & & &  \\
\hline %
$x_1$ & - &  & & & & & & &  \\
\text{m} & $x_2$ &  & & & & & & &  \\
$x_3$ & - &  & & & & & & &  \\
\text{m} & $x_4$  &  & & & & & & &  \\
\hline %
$x_5$ & $x_6$ &  & & & & & & &  \\
$x_6$ & $x_7$ &  & & & & & & &  \\
$x_7$ & $x_8$ &  & & & & & & &  \\
$x_8$ & $x_9$  &  & & & & & & &  \\
\hline %
\text{m} & $x_5$ &  & & & & & & &  \\
\text{m} & $x_6$ &  & & & & & & &  \\
$x_7$ & - &  & & & & & & &  \\
\text{m} & $x_8$  &  & & & & & & &  \\
\hline %
\end{tblr}
}
\caption{Training a hybrid model for block of size $k=4$. We show input tokens (left column), target tokens used for Cross-Entropy loss (second columns) and attention matrix.  }\label{fig:training}
\end{figure}